\documentclass{article} 
\usepackage{times}

\usepackage{amsmath,amsfonts,bm}

\def\eqref#1{equation~\ref{#1}}

\def\1{\bm{1}}

\DeclareMathAlphabet{\mathsfit}{\encodingdefault}{\sfdefault}{m}{sl}
\SetMathAlphabet{\mathsfit}{bold}{\encodingdefault}{\sfdefault}{bx}{n}

\usepackage[letterpaper,top=2cm,bottom=2cm,left=3cm,right=3cm,marginparwidth=1.75cm]{geometry}

\usepackage{hyperref}
\usepackage{url}
\usepackage{booktabs}
\usepackage{placeins}
\usepackage{multirow}
\usepackage{graphicx}
\usepackage{natbib}
\usepackage{tabularx}
\usepackage{float}
\usepackage{stfloats}
\usepackage{amssymb}
\usepackage[ruled,vlined]{algorithm2e}

\newtheorem{theorem}{Theorem}
\newtheorem{proof}{Proof}
\newtheorem{corollary}{Corollary}
\newtheorem{assumption}{Assumption}

\title{Learning Robust Recommenders from Noisy Implicit Feedback
via GMM-Weighted Bayesian Transition Matrix}

\author{
  Zongyu Li\thanks{\href{mailto:zongyuli@bjtu.edu.cn}{zongyuli@bjtu.edu.cn}}%
    \thanks{\textsuperscript{*}Corresponding author}\\
  Guangdong University of Technology, Guangzhou, China
  \and
  Xuanyu Liu \href{mailto:liuxuanyu23@mails.ucas.edu.cn}{liuxuanyu23@mails.ucas.edu.cn}\\
  University of Chinese Academy of Sciences, Beijing, China
  \and
  Gongce Cao \href{mailto:caogongce25@mails.ucas.ac.cn}{caogongce25@mails.ucas.ac.cn}\\
  University of Chinese Academy of Sciences, Beijing, China
  \and
  Shirui Sun \href{mailto:shiruisun910@gmail.com}{shiruisun910@gmail.com}\\
  Capital Normal University, Dalian, China
  \and
  Yaqi Fang \href{mailto:fangyaqi@stu.xmu.edu.cn}{fangyaqi@stu.xmu.edu.cn}\\
  Xiamen University, Xiamen, China
  \and
  Yongshuai Yu \href{mailto:25125355@bjtu.edu.cn}{25125355@bjtu.edu.cn}\\
  Beijing Jiaotong University, Beijing, China
}

\begin{document}
\raggedbottom

\maketitle

\begin{abstract}
Label noise is a central challenge in learning from implicit feedback for recommendation. Conventional approaches discard noisy examples for robustness, but this sacrifices data efficiency. Unlike filtering approaches, Bayes-label transition matrix (BLTM) based methods keep all data, but their transition matrix estimates are skewed in practice. To reduce this skew, we introduce GMM-weighted Bayes-label Transition Matrix (RGBT), which augments BLTM with GMM-based instance weights. A GMM assigns each instance a reliability score, and these scores calibrate the BLTM to reduce bias. We show theoretically that RGBT retains all samples, yields consistent estimates, and provably reduces variance compared to CLTM, generalizing to any method with deterministic labels and non-negative weights. Experiments on real and synthetic datasets show that RGBT handles noisy samples more effectively than sample-selection methods, and calibrates the transition matrix more accurately than existing approaches.
\end{abstract}

\section{Introduction}

Recommender systems have advanced considerably in recent years \cite{Recommender_survey_1,Recommender_survey_2,Recommender_survey_3,Recommender_survey_4}. Most of these systems rely on implicit feedback clicks, purchases, and similar signals to infer user preferences, assuming these behaviors reflect interests \cite{Recommender_noise_1,Recommender_noise_2,Recommder_noise_3,Recommder_noise_4,Recommder_noise_5,Recommder_noise_6}. However, notable exceptions exist. For instance, a user might purchase an item only to return it later, or click on a video but exit almost immediately. In such cases, the interactions do not reflect genuine user interest, and are often referred to as noisy interactions in the literature \cite{ADT,DeCA,SGDL,BOD,DCF}.

Existing denoising methods typically address noisy implicit feedback by directly discarding unreliable samples to ensure robustness, yet this strategy inevitably leads to suboptimal data utilization \cite{ADT,DeCA,SGDL,BOD,DCF}. For instance, ADT \cite{ADT} adaptively prunes or down-weights high-loss interactions; DeCA \cite{DeCA} aligns model predictions while maximizing data likelihood by leveraging cross-model agreement; SGDL \cite{SGDL} employs a self-guided meta-learning strategy that collects reliable signals early in training to guide subsequent denoising; BOD \cite{BOD} formulates the denoising task as a bi-level optimization problem efficiently solved via a weight generator; and DCF \cite{DCF} implements a dual-correction mechanism through stabilized sample dropping and progressive label refinement. While these methods have demonstrated certain effectiveness in practice, they either fail to adequately address inherent biases in noise distribution modeling or underutilize the informational value present in noisy samples. More critically, the majority of these approaches are heuristic in nature, lacking the theoretical foundations to guarantee that the denoised model converges to the optimal cleaner. 

Estimating the noise transition matrix is a key challenge in learning from noisy implicit feedback. BLTM-based methods \cite{BLTM} can use all data in theory, but their estimates are often biased in practice. Prior work has explored various strategies, including consistency modeling \cite{RRFN}, matrix factorization \cite{Dual-T}, anchor-free estimation \cite{T-Revision}, volume minimization \cite{VolMinNet}, cycle-consistency regularization \cite{CCR}, instance-dependent estimation \cite{IDNT}, and self-supervised learning \cite{CoNL}. Yet none work well with complex real-world noise. The resulting bias can then hurt the final recommendation model.

To address these issues, we propose RGBT, which integrates GMM-based weighting with transition matrix correction. Our main contributions are as follows.

1) We develop a robust recommendation framework that effectively leverages noisy samples through Bayesian label transition matrix (BLTM) modeling, while achieving more accurate bias correction in transition matrix estimation via Gaussian Mixture Model (GMM)-based calibration.

2) We prove theoretically that RGBT uses all data, yields consistent estimates, and has lower variance than existing transition-matrix methods, extending beyond GMM-based calibration to any method with deterministic labels and non-negative weights.

3) Experiments on real and synthetic datasets show that RGBT handles noise better than sample-selection baselines and calibrates the transition matrix more accurately than existing approaches. 

\section{Preliminaries}

Consider a recommendation system with $m$ users and $n$ items. Let $Y \in \{1,\dots,K\}^{m \times n}$ be the clean user–item rating matrix, where $y_{ui}$ is the true rating of user $u$ for item $i$. In practice, we only observe a noisy version $\widetilde Y \in \{1,\dots,K\}^{m \times n}$, in which each entry is flipped from its true value to another rating with probabilities that depend on the instance.

We learn user embeddings $\mathbf{p}_u \in \mathbb{R}^d$ and item embeddings $\mathbf{q}_i \in \mathbb{R}^d$. For each user-item pair $(u,i)$, we form a feature vector $\mathbf{x}_{ui} = [\mathbf{p}_u;\mathbf{q}_i] \in \mathbb{R}^{2d}$. The observed label $\tilde{y}_{ui} \in \{1,\dots,K\}$ is noisy. This yields a $K$-class classification problem with noisy labels.

For robustness, we keep only high-confidence predictions as distilled samples. Let $\hat{\boldsymbol{\eta}}(\mathbf{x}_{ui}) \in [0,1]^K$ be the estimated class probability vector. Given $\rho_{\text{max}} \in [0,1]$ as the maximum noise rate threshold, the distilled label is defined as:
\begin{equation}
y^*_{ui} = 
\begin{cases}
\arg\max_{k} \hat{\eta}_k(\mathbf{x}_{ui}) & \text{if } \max_k \hat{\eta}_k(\mathbf{x}_{ui}) > \frac{1 + \rho_{\text{max}}}{2}, \\
\varnothing & \text{otherwise}.
\end{cases}
\label{eq:distilled_label}
\end{equation}
The resulting set $\mathcal{D}_{\text{distilled}} = \{(\mathbf{x}_i, \tilde{y}_i, y^*_i)\}_{i=1}^M$ keeps only predictions above this threshold, where $M$ is the number of retained instances and $y^*_i$ is the inferred Bayes optimal label.

Under the Instance-Dependent Noise (IDN) assumption, the noise transition matrix depends on both the true label and the instance features. We define $\mathbf{T}(\mathbf{x}) \in [0,1]^{K\times K}$ with entries
\begin{equation}
T_{ij}(\mathbf{x}) = \mathbb{P}(\widetilde{Y} = j \mid Y = i, \mathbf{X} = \mathbf{x}), \quad i,j \in \{1,\dots,K\},
\label{eq:transition_matrix}
\end{equation}
where for each $i$ and $\mathbf{x}$, $T_{ij}(\mathbf{x}) \ge 0$ and $\sum_{j=1}^K T_{ij}(\mathbf{x}) = 1$.

With the distilled samples, we define the Bayes-label transition matrix (BLTM) under IDN as
\begin{equation}
T^*_{ij}(\mathbf{x}) = \mathbb{P}(\widetilde{Y} = j \mid Y^* = i, \mathbf{X} = \mathbf{x}),
\label{eq:bltm_definition}
\end{equation}
where $Y^* = \arg\max_{y} \mathbb{P}(Y = y \mid \mathbf{x})$ is the Bayes optimal label. This matrix is estimated by a transition network with parameters $\theta$:
\begin{equation}
\hat{T}^*_{ij}(\mathbf{x};\theta) = \mathbb{P}(\widetilde{Y} = j \mid Y^* = i, \mathbf{X} = \mathbf{x}; \theta), \quad \mathbf{x} \in \mathcal{D}_{\text{distilled}}.
\label{eq:bltm_network}
\end{equation}

A GMM is used to assess reliability. With $C$ components and parameters $\Theta = \{(\pi_c, \boldsymbol{\mu}_c, \boldsymbol{\Sigma}_c)\}_{c=1}^C$, the weight for a feature vector $\mathbf{x}$ is
\begin{equation}
w(\mathbf{x}) = \frac{\pi_r \mathcal{N}(\mathbf{x}|\boldsymbol{\mu}_r,\boldsymbol{\Sigma}_r)}{\sum_{c=1}^C \pi_c \mathcal{N}(\mathbf{x}|\boldsymbol{\mu}_c,\boldsymbol{\Sigma}_c)},
\label{eq:reliability_weight}
\end{equation}
where component $r$ is selected as the reliable one (e.g., with the largest $\|\boldsymbol{\mu}_c\|^2$). These weights then calibrate the transition matrix.

We learn a classifier $f: \mathcal{X} \to \{1,\dots,K\}$ from noisy data. The goal is to minimize the expected risk $\mathbb{E}[\mathbb{I}(f(\mathbf{X}) \neq Y)]$ with respect to the clean distribution, using distilled samples and reliability-weighted BLTM estimation.

\section{Method}

\subsection{Robust GMM-Weighted Bayes-label Transition Framework}

Label noise makes it hard to learn from implicit feedback in recommender systems. RGBT handles this with three components: BLTM for data utilization via distilled samples, GMM for reliability, and GMM-calibrated transition matrix estimation.

\subsubsection{BLTM with Distilled Samples for Enhanced Data Utilization}

Traditional denoising methods discard noisy instances and sacrifice data efficiency. We avoid this with the Bayes-label transition matrix (BLTM) using distilled samples. Distilled examples $(\mathbf{x},\tilde{y}, y^{*})$ are obtained by keeping high-confidence predictions. Let $\hat{\boldsymbol{\eta}}(\mathbf{x}) \in [0,1]^K$ be the estimated class probability vector and $\rho_{\text{max}}$ the maximum noise rate threshold. The distilled label is
\begin{equation}
y^{*} = 
\begin{cases}
\arg\max_{k} \hat{\eta}_k(\mathbf{x}) & \text{if } \max_k \hat{\eta}_k(\mathbf{x}) > \frac{1 + \rho_{\text{max}}}{2}, \\
\varnothing & \text{otherwise}.
\end{cases}
\label{eq:distilled_label_method}
\end{equation}
Keeping only predictions above this threshold yields $\mathcal{D}_{\text{distilled}} = \{(\mathbf{x}_i, \tilde{y}_i, y^{*}_i)\}_{i=1}^M$.

Under the instance-dependent noise assumption, the BLTM is defined as:
\begin{equation}
T^*_{ij}(\mathbf{x}) = \mathbb{P}(\tilde{Y} = j \mid Y^* = i, \mathbf{X} = \mathbf{x}), \quad i,j \in \{1,\dots,K\},
\label{eq:bltm_definition_method}
\end{equation}
where $Y^* = \arg\max_{y} \mathbb{P}(Y = y \mid \mathbf{x})$ is the Bayes optimal label, $\tilde{Y}$ is the noisy label, and $\mathbf{X}$ is the feature variable.

We train a transition network parameterized by $\theta$ to estimate the BLTM using distilled samples:
\begin{equation}
\hat{T}^*_{ij}(\mathbf{x};\theta) = \mathbb{P}(\tilde{Y} = j \mid Y^* = i, \mathbf{X} = \mathbf{x}; \theta) \quad \text{for } \mathbf{x} \in \mathcal{D}_{\text{distilled}}.
\label{eq:bltm_network_method}
\end{equation}
Here $\hat{T}^*(\mathbf{x};\theta) \in [0,1]^{K\times K}$ is the estimated BLTM, where each row sums to 1.

The base learning objective with distilled samples is:
\begin{equation}
\mathcal{L}_{\text{BLTM}}(\theta) = -\frac{1}{M}\sum_{(\mathbf{x},\tilde{y},y^{*})\in\mathcal{D}_{\text{distilled}}} \tilde{\boldsymbol{y}}^{\top} \log\left(\boldsymbol{y}^{*\top} \hat{T}^*(\mathbf{x};\theta)\right),
\label{eq:bltm_loss}
\end{equation}
where $\mathcal{L}_{\text{BLTM}}$ denotes the BLTM loss, $\tilde{\boldsymbol{y}}$ is the one-hot encoding of the observed noisy label $\tilde{y}$, and $\boldsymbol{y}^{*}$ is the one-hot encoding of the distilled Bayes optimal label $y^{*}$.

\subsubsection{GMM-based Reliability Extraction}

BLTM improves data utilization via distilled samples, but sample quality varies. A GMM provides reliability weights for calibration.

For each user-item pair $(u, v)$ in the distilled set $\mathcal{D}_{\text{distilled}}$, we compute the prediction score $\hat{y}_{uv}$ and construct reliability features. Let $\mathbf{p}_u \in \mathbb{R}^d$ and $\mathbf{q}_v \in \mathbb{R}^d$ be user and item embeddings respectively, with $d$ denoting the embedding dimension. The prediction score is:
\begin{equation}
\hat{y}_{uv} = \mathbf{p}_u^{\top}\mathbf{q}_v.
\label{eq:prediction_score}
\end{equation}

The co-occurrence feature $c_{uv}$ captures neighborhood similarity:
\begin{equation}
c_{uv} = \sum_{u^{\prime}\in\mathcal{U}_v}\mathbb{I}(|\mathcal{I}_u\cap\mathcal{I}_{u^{\prime}}| > 1),
\label{eq:cooccurrence}
\end{equation}
where $\mathcal{U}_v = \{u: (u,v) \in \mathcal{D}_{\text{train}}\}$ is the set of users who interacted with item $v$, $\mathcal{I}_u = \{v: (u,v) \in \mathcal{D}_{\text{train}}\}$ is the item set of user $u$, and $\mathbb{I}(\cdot)$ is the indicator function.

The reliability feature vector is then $\mathbf{x}_{uv} = (\hat{y}_{uv}, c_{uv}) \in \mathbb{R}^2$.

A two-component Gaussian Mixture Model (GMM) models the distribution of reliability features:
\begin{equation}
p(\mathbf{x}|\Theta) = \sum_{k=1}^{2}\pi_k\mathcal{N}(\mathbf{x}|\boldsymbol{\mu}_k,\boldsymbol{\Sigma}_k),
\label{eq:gmm_distribution}
\end{equation}
where $\Theta = \{\pi_k, \boldsymbol{\mu}_k, \boldsymbol{\Sigma}_k\}_{k=1}^2$ are GMM parameters: $\pi_k \in [0,1]$ (mixing coefficients with $\sum_k \pi_k = 1$), $\boldsymbol{\mu}_k \in \mathbb{R}^2$ (mean vectors), and $\boldsymbol{\Sigma}_k \in \mathbb{R}^{2\times2}$ (covariance matrices).

From the fitted GMM, we derive instance-specific reliability weights for distilled samples:
\begin{equation}
w_{uv} = \frac{\pi_{k^*}\mathcal{N}(\mathbf{x}_{uv}|\boldsymbol{\mu}_{k^*},\boldsymbol{\Sigma}_{k^*})}{\sum_{k=1}^{2}\pi_k\mathcal{N}(\mathbf{x}_{uv}|\boldsymbol{\mu}_k,\boldsymbol{\Sigma}_k)},
\label{eq:gmm_weight}
\end{equation}
where $k^* = \arg\max_{k\in\{1,2\}}\|\boldsymbol{\mu}_k\|^2$ identifies the component with larger norm (assumed to represent reliable samples), and $w_{uv} \in [0,1]$ is the reliability weight for the user-item pair $(u,v)$.

\subsubsection{GMM-Calibrated BLTM Estimation and Classification}

The final component integrates reliability weights to calibrate BLTM estimation and enhance classification. We reformulate the transition matrix learning with reliability-weighted distilled samples:
\begin{equation}
\mathcal{L}_{\text{calibrated}}(\mathbf{w}) = -\frac{1}{M}\sum_{(\mathbf{x},\tilde{y},y^{*})\in\mathcal{D}_{\text{distilled}}} w(\mathbf{x}) \cdot \boldsymbol{y}^{*\top} \log\left(f(\mathbf{x};\mathbf{w})\right),
\label{eq:calibrated_loss}
\end{equation}
where $\mathcal{L}_{\text{calibrated}}(\mathbf{w})$ is the calibrated loss, $M = |\mathcal{D}_{\text{distilled}}|$ is the number of distilled samples, $w(\mathbf{x}) \in [0,1]$ is the reliability weight from GMM, $\boldsymbol{y}^{*} \in \{0,1\}^K$ is the one-hot encoding of the distilled Bayes optimal label $y^{*}$, and $f(\mathbf{x};\mathbf{w}): \mathbb{R}^p \to \mathbb{R}^K$ is the classifier that outputs the estimated clean class posterior distribution.

For the final classification, we employ the calibrated BLTM for noise correction. The classifier $f(\mathbf{x};\mathbf{w})$ with parameters $\mathbf{w}$ outputs the estimated clean class posterior distribution. The classification loss is:
\begin{equation}
\mathcal{L}_{\text{class}}(\mathbf{w}, \theta) = -\frac{1}{N}\sum_{(\mathbf{x},\tilde{y})\in\mathcal{D}} \tilde{\boldsymbol{y}}^{\top} \log\left(\hat{T}^*(\mathbf{x};\theta) \, f(\mathbf{x};\mathbf{w})\right),
\label{eq:class_loss}
\end{equation}
where $\mathcal{L}_{\text{class}}(\mathbf{w}, \theta)$ is the classification loss, $N = |\mathcal{D}|$ is the total number of training samples, $\mathcal{D}$ is the complete training dataset, $\tilde{\boldsymbol{y}} \in \{0,1\}^K$ is the one-hot encoding of the observed noisy label $\tilde{y}$, and the product $\hat{T}^*(\mathbf{x};\theta) \, f(\mathbf{x};\mathbf{w}) \in \mathbb{R}^K$ yields the predicted noisy class distribution.

The complete RGBT framework jointly optimizes:
\begin{equation}
\mathcal{L}_{\text{RGBT}}(\mathbf{w},\theta) = \mathcal{L}_{\text{class}}(\mathbf{w}, \theta) + \lambda \mathcal{L}_{\text{calibrated}}(\mathbf{w}),
\label{eq:rgbt_loss}
\end{equation}
where $\lambda > 0$ is a balancing hyperparameter that controls the relative weight of the classification and transition matrix learning objectives.

We summarize the RGBT training procedure in Algorithm~\ref{alg:rgbt}.

\subsection{Theoretical Analysis}
\label{sec:theory}

\subsubsection{Noise Model and Bayes Optimality}

We work with bounded instance-dependent label noise. Let $\eta_y(\mathbf{x}) = \mathbb{P}(Y = y \mid \mathbf{X} = \mathbf{x})$ be the clean class-posterior, and $\tilde{\eta}_y(\mathbf{x}) = \mathbb{P}(\widetilde{Y} = y \mid \mathbf{X} = \mathbf{x})$ the noisy version. The relation between the two is governed by a transition matrix $T_{y'y}(\mathbf{x}) = \mathbb{P}(\widetilde{Y} = y \mid Y = y', \mathbf{X} = \mathbf{x})$:
\begin{equation}
\tilde{\eta}_y(\mathbf{x}) = \sum_{y' \in \mathcal{Y}} T_{y'y}(\mathbf{x}) \eta_{y'}(\mathbf{x}).
\label{eq:noise_transition_main}
\end{equation}

Under this model, the Bayes optimal classifier is unchanged by the noise. We state this formally below.

\begin{theorem}[Bayes optimality under noise]\label{thm:bayes_optimal}
For any $\mathbf{x} \in \operatorname{supp}(P_D)$,
$g^*_{D^*}(\mathbf{x}) = g^*_D(\mathbf{x}) = \arg\max_y P_D(Y=y|\mathbf{x})$.
\end{theorem}

The estimation and correction steps below require the transition matrix
to be estimated stably. With a single noisy label per instance,
$T^{*}(\mathbf{x})$ is not identifiable from the data alone; we
therefore adopt the manifold-regularization assumption and defer the
formal statement to Appendix~\ref{app:manifold}.

\begin{algorithm}[t]
\caption{Robust GMM-Weighted Bayes-Label Transition Training}
\label{alg:rgbt}
\SetAlgoLined
\KwIn{Noisy dataset $\mathcal{D}$, noise rate $\rho_{\max}$, learning rates $\eta_{\mathbf{w}},\eta_{\theta}$, refresh interval $t_{\text{ref}}$}
\KwOut{Classifier $f(\cdot;\mathbf{w})$, transition estimator $\hat{T}^*(\cdot;\theta)$}
Initialize $\mathbf{w},\theta$; $\mathcal{D}_{\text{dist}}\leftarrow\emptyset$; $t\leftarrow0$\;
\While{not converged}{
    \If{$t\bmod t_{\text{ref}}=0$ or $\mathcal{D}_{\text{dist}}=\emptyset$}{
        Extract distilled examples $(\mathbf{x},\tilde{y},y^*)$ from $\mathcal{D}$ where $\max f(\mathbf{x};\mathbf{w})>(1+\rho_{\max})/2$\;
        Fit GMM and compute reliability weights $w(\mathbf{x})$ on $\mathcal{D}_{\text{dist}}$\;
    }
    Update $\mathbf{w}$ with $\mathcal{L}_{\text{calibrated}}$ using $\mathcal{D}_{\text{dist}}$ weighted by $w(\mathbf{x})$\;
    Update $\theta$ and $\mathbf{w}$ with $\mathcal{L}_{\text{class}}$ on $\mathcal{D}$\;
    $t\leftarrow t+1$\;
}
\Return $f(\cdot;\mathbf{w})$, $\hat{T}^*(\cdot;\theta)$
\end{algorithm}

\subsubsection{Distillation: Extracting Reliable Labels}

Directly estimating the transition matrix from noisy data is difficult because the clean labels are unobserved. To get around this, we first extract a set of high-confidence examples.

We keep predictions that pass a confidence threshold as distilled labels:
\begin{equation}
y^* = 
\begin{cases}
\arg\max_k f_k(\mathbf{x};\mathbf{w}) & \text{if } \max_k f_k(\mathbf{x};\mathbf{w}) > \frac{1+\rho_{\max}}{2}, \\
\varnothing & \text{otherwise}.
\end{cases}
\label{eq:distill_main}
\end{equation}
The threshold is chosen so that any retained prediction is better than random, as stated below.

\begin{theorem}[Distillation condition]\label{thm:distill_cond}
If $\max_k f_k(\mathbf{x};\mathbf{w}) > (1+\rho_{\max})/2$, then $\eta_{y^*}(\mathbf{x}) > 1/2$.
\end{theorem}
The proof follows from the noise bound $\rho_{\max} < 1/2$ and is deferred to Appendix~\ref{app:theory}.

These distilled labels are not all equally reliable. We use a GMM to assign each sample a weight $w(\mathbf{x})$, which reflects its estimated reliability. The calibrated learning objective is then
\begin{equation}
\mathcal{L}_{\text{calibrated}}(\mathbf{w}) = -\frac{1}{M}\sum_{(\mathbf{x},\tilde{y},y^*)\in\mathcal{D}_{\text{distilled}}} w(\mathbf{x}) \, \boldsymbol{y}^{*\top} \log f(\mathbf{x};\mathbf{w}).
\label{eq:cal_loss_main}
\end{equation}

Minimizing this loss drives the classifier toward the clean posterior, as the next theorem shows.

\begin{theorem}[Classifier consistency]\label{thm:classifier_consistency}
As $M \to \infty$ and $\rho_{\max} \to 0$, the minimizer of $\mathcal{L}_{\text{calibrated}}$ in \eqref{eq:cal_loss_main} satisfies
$f(\mathbf{x};\hat{\mathbf{w}}) \xrightarrow{p} \eta(\mathbf{x})$.
\end{theorem}

\subsubsection{Learning the Transition Matrix}

With the classifier approaching $\eta(\mathbf{x})$, we can estimate the transition matrix. We do this by minimizing
\begin{equation}
\mathcal{L}_{\text{class}}(\mathbf{w},\theta) = -\frac{1}{N}\sum_{(\mathbf{x},\tilde{y})\in\mathcal{D}} \tilde{\boldsymbol{y}}^{\top} \log\left(\hat{T}^*(\mathbf{x};\theta) f(\mathbf{x};\mathbf{w})\right).
\label{eq:cls_loss_main}
\end{equation}
The intuition is straightforward: $\hat{T}^* f$ should match the observed noisy posterior $\tilde{\eta}$. Since $f \to \eta$ and $\tilde{\eta} = T^* \eta$, consistency follows.

\begin{theorem}[Transition matrix consistency]\label{thm:transition_consistency}
Under $\mathcal{L}_{\text{RGBT}} = \mathcal{L}_{\text{class}} + \lambda \mathcal{L}_{\text{calibrated}}$,
$\hat{T}^*(\mathbf{x};\hat{\theta}) \xrightarrow{p} T^*(\mathbf{x})$.
\end{theorem}

\subsubsection{Variance Reduction}

We use distilled labels instead of the plug-in estimates from a
noisy-label classifier, as conventional CLTM does. The difference is
variance: distilled labels are deterministic given \(\mathbf{x}\), whereas
plug-in estimates are random.

\begin{theorem}[Variance reduction]\label{thm:variance_reduction}
Under bounded noise with $\rho_{\max} < 1/2$,
\[
\lim_{n\to\infty} n\cdot \operatorname{Var}[\hat{T}^{\text{BLTM}}_{ij}(\mathbf{x})]
\leq
\lim_{n\to\infty} n\cdot \operatorname{Var}[\hat{T}^{\text{CLTM}}_{ij}(\mathbf{x})].
\]
\end{theorem}

Appendix~\ref{app:theory} provides the full derivations, the proof of
Theorem~\ref{thm:variance_reduction}, a finite-sample bound
(Corollary~\ref{cor:finite_sample}), and a general condition extending the
result beyond GMM weighting (Corollary~\ref{cor:general_variance}).

\raggedbottom
\begingroup
\setlength{\parskip}{2pt}
\setlength{\textfloatsep}{6pt}
\setlength{\intextsep}{6pt}
\setlength{\floatsep}{6pt}
\section{Experiments}

We evaluate RGBT on six recommendation backbones and three real-world datasets to answer four questions. RQ1: How does RGBT compare with standard training and denoising methods? RQ2: How accurately does our calibration method estimate transition matrices under synthetic flip noise compared with existing matrix-based approaches, and how does this accuracy translate to debiasing performance? RQ3: How sensitive is RGBT to its calibration and distillation hyperparameters? RQ4: How do reliability weighting and transition correction contribute to recommendation performance?

\subsection{Experimental Settings}

\subsubsection{Datasets}
We use three public datasets: Adressa for news reading, MovieLens for movie ratings, and Yelp for business recommendation~\citep{SGDL,ADT,DCF,DeCA}. Following the DCF preprocessing protocol~\citep{DCF}, the recommendation test sets retain interactions with dwell times of at least 10 seconds for Adressa, ratings of five for MovieLens, and ratings higher than three for Yelp. The training and validation sets remain noisy and unfiltered.

\subsubsection{Noise Simulation for RQ2}
\label{sec:noise_simulation}
Following RRFN~\citep{RRFN}, we inject two types of rating flip noise into clean MovieLens data. For a clean rating $y$ among $K$ classes, symmetric flip noise assigns the noisy label $\tilde{y}=k$ with probability $\eta/(K-1)$ for each $k\neq y$, whereas pairwise flip noise assigns $\tilde{y}=(y+1)\bmod K$ with probability $\eta$, using zero-based class indices. In both settings, the rating remains unchanged with probability $1-\eta$. We use noise rates $\eta\in\{0.1,0.2,0.3,0.4\}$.

\subsubsection{Evaluation protocols}

Following the protocol established in prior research, we use the
training, validation, and clean test splits with an 8:1:1
ratio for RQ1 and RQ3 ~\citep{ADT,DeCA,SGDL,DCF}.
For performance evaluation, we adopt the two widely used ranking
metrics common in denoising recommendation literature: NDCG@K and
Recall@K. Higher scores on both metrics indicate better model
performance. Table~\ref{tab:overall_performance} reports results
for $K=10$ and $K=50$ across the three datasets.

To quantify noise estimation accuracy in RQ2, we average the estimated instance-dependent transition matrices over the validation interactions and compute the entrywise $L_1$ discrepancy from the known transition matrix ~\citep{MEIDTM,CoNL}.

\subsubsection{Baselines}

We evaluate RGBT on six backbones: GMF and NeuMF~\citep{GMF+NeuMF}, NGCF~\citep{NGCF}, LightGCN~\citep{LightGCN}, XSimGCL~\citep{XSimGCL}, and CDAE~\citep{CDAE}. For recommendation performance, we compare against Normal (BCE training), WBPR~\citep{WBPR}, WRMF~\citep{WRMF}, T-CE~\citep{ADT}, DeCA~\citep{DeCA}, and SGDL~\citep{SGDL}. We adapt TCL~\citep{TCL} and C2D~\citep{C2D} to recommendation. For transition matrix estimation in RQ2, we additionally consider CCR~\citep{CCR}, Dual-T~\citep{Dual-T}, T-Revision~\citep{T-Revision}, VolMinNet~\citep{VolMinNet}, RRFN~\citep{RRFN}, and CONL~\citep{CoNL}.

\vspace{-5pt}
\subsection{Performance Comparison (RQ1)}

To answer RQ1, Table~\ref{tab:overall_performance} compares RGBT with baseline models across datasets; additional WRMF, T-CE, and BLTM results are in Appendix~\ref{app:baseline_results}. On MovieLens and Yelp, RGBT improves over Normal on all four reported metrics for all six backbones. On Adressa, the gains depend on the backbone and metric: NGCF and LightGCN show substantial improvements in Recall@50, while GMF does not consistently outperform the strongest baselines.

\begin{table}[!htb]
\centering
\caption{Performance comparison on six backbones and three datasets. Higher is better. Bold and underlining indicate the best and second-best scores among the methods shown, including ties.}
\label{tab:overall_performance}
\fontsize{8}{8.9}\selectfont
\setlength{\tabcolsep}{1.9pt}
\renewcommand{\arraystretch}{1.03}
\begin{tabular*}{\linewidth}{@{\extracolsep{\fill}}ll|rrrr|rrrr|rrrr@{}}
\toprule
\multicolumn{2}{c|}{Dataset} & \multicolumn{4}{c|}{Adressa} & \multicolumn{4}{c|}{MovieLens} & \multicolumn{4}{c}{Yelp} \\
\midrule
Backbone & Method & R@10 & R@50 & N@10 & N@50 & R@10 & R@50 & N@10 & N@50 & R@10 & R@50 & N@10 & N@50 \\
\midrule
\multirow{7}{*}{GMF}  &  Normal  & \textbf{0.1881} & 0.2544 & 0.1137 & 0.1314 & 0.0769 & 0.1987 & 0.0595 & 0.0963 & 0.0261 & 0.0859 & 0.0193 & 0.0362 \\
 &  WBPR  & 0.1708 & 0.2418 & 0.1019 & 0.1209 & 0.0696 & 0.1918 & 0.0552 & 0.0921 & 0.0205 & 0.0708 & 0.0148 & 0.0290 \\
 &  DeCA  & 0.1511 & 0.2541 & 0.0976 & 0.1263 & 0.0765 & 0.1837 & 0.0568 & 0.0901 & 0.0241 & 0.0759 & 0.0179 & 0.0327 \\
 &  SGDL  & \underline{0.1870} & \textbf{0.2653} & \textbf{0.1139} & \textbf{0.1347} & \underline{0.0779} & \underline{0.1991} & \underline{0.0654} & \underline{0.1021} & 0.0227 & 0.0754 & 0.0166 & 0.0316 \\
 &  TCL  & 0.1788 & 0.2556 & 0.1089 & 0.1293 & 0.0399 & 0.1663 & 0.0347 & 0.0722 & \underline{0.0283} & 0.0890 & \underline{0.0205} & 0.0377 \\
 &  C2D  & 0.1639 & \underline{0.2612} & 0.1043 & 0.1301 & 0.0759 & 0.1835 & 0.0574 & 0.0907 & 0.0282 & \underline{0.0911} & \underline{0.0205} & \underline{0.0383} \\
\cmidrule(lr){2-14}
 &  \textbf{RGBT (Ours)}  & 0.1866 & 0.2546 & \underline{0.1138} & \underline{0.1320} & \textbf{0.0836} & \textbf{0.2124} & \textbf{0.0664} & \textbf{0.1069} & \textbf{0.0302} & \textbf{0.1068} & \textbf{0.0210} & \textbf{0.0427} \\
\midrule
\multirow{7}{*}{NeuMF}  &  Normal  & 0.2856 & 0.3628 & \textbf{0.1790} & \textbf{0.1995} & 0.0644 & 0.1892 & 0.0554 & 0.0927 & 0.0231 & 0.0773 & 0.0163 & 0.0317 \\
 &  WBPR  & 0.1969 & \underline{0.3654} & 0.1300 & 0.1769 & 0.0665 & 0.1996 & 0.0538 & 0.0937 & 0.0185 & 0.0697 & 0.0130 & 0.0274 \\
 &  DeCA  & 0.2366 & 0.3539 & 0.1537 & 0.1864 & 0.0516 & 0.1605 & 0.0416 & 0.0757 & 0.0205 & 0.0714 & 0.0147 & 0.0290 \\
 &  SGDL  & 0.1805 & 0.2602 & 0.1107 & 0.1317 & \underline{0.0745} & \underline{0.2033} & \underline{0.0602} & \underline{0.0987} & \underline{0.0270} & \underline{0.0832} & \textbf{0.0200} & \underline{0.0359} \\
 &  TCL  & 0.2832 & 0.3611 & 0.1734 & 0.1943 & 0.0548 & 0.1843 & 0.0445 & 0.0852 & 0.0164 & 0.0538 & 0.0113 & 0.0220 \\
 &  C2D  & \underline{0.2865} & 0.3630 & 0.1749 & 0.1954 & 0.0725 & 0.1859 & 0.0585 & 0.0925 & 0.0157 & 0.0535 & 0.0112 & 0.0220 \\
\cmidrule(lr){2-14}
 &  \textbf{RGBT (Ours)}  & \textbf{0.2890} & \textbf{0.3686} & \underline{0.1756} & \underline{0.1964} & \textbf{0.0879} & \textbf{0.2151} & \textbf{0.0745} & \textbf{0.1130} & \textbf{0.0272} & \textbf{0.0987} & \underline{0.0192} & \textbf{0.0394} \\
\midrule
\multirow{7}{*}{NGCF}  &  Normal  & 0.1979 & 0.2603 & 0.1215 & 0.1384 & 0.0646 & 0.1864 & 0.0525 & 0.0886 & \underline{0.0320} & \underline{0.0958} & \underline{0.0229} & \underline{0.0410} \\
 &  WBPR  & 0.1799 & 0.2458 & 0.1109 & 0.1285 & 0.0688 & 0.1880 & 0.0538 & 0.0906 & 0.0303 & 0.0930 & 0.0219 & 0.0395 \\
 &  DeCA  & 0.1747 & 0.2328 & 0.1057 & 0.1214 & 0.0403 & 0.1044 & 0.0390 & 0.0587 & 0.0212 & 0.0664 & 0.0156 & 0.0285 \\
 &  SGDL  & 0.1924 & 0.2546 & 0.1137 & 0.1305 & 0.0769 & \underline{0.1975} & 0.0590 & 0.0949 & 0.0272 & 0.0858 & 0.0195 & 0.0360 \\
 &  TCL  & 0.2738 & 0.3582 & \underline{0.1694} & \underline{0.1933} & 0.0797 & 0.1936 & \underline{0.0645} & \underline{0.0997} & 0.0278 & 0.0871 & 0.0203 & 0.0369 \\
 &  C2D  & \underline{0.2749} & \underline{0.3588} & 0.1688 & 0.1922 & \underline{0.0800} & 0.1902 & 0.0638 & 0.0975 & 0.0235 & 0.0736 & 0.0172 & 0.0314 \\
\cmidrule(lr){2-14}
 &  \textbf{RGBT (Ours)}  & \textbf{0.2755} & \textbf{0.3680} & \textbf{0.1711} & \textbf{0.1960} & \textbf{0.0847} & \textbf{0.2097} & \textbf{0.0733} & \textbf{0.1118} & \textbf{0.0328} & \textbf{0.1069} & \textbf{0.0233} & \textbf{0.0444} \\
\midrule
\multirow{7}{*}{LightGCN}  &  Normal  & 0.1971 & \underline{0.2604} & 0.1204 & 0.1377 & 0.0648 & 0.1857 & 0.0530 & 0.0885 & 0.0321 & 0.0958 & 0.0231 & 0.0411 \\
 &  WBPR  & 0.1898 & 0.2543 & 0.1164 & 0.1339 & 0.0673 & 0.1911 & 0.0530 & 0.0906 & 0.0306 & 0.0935 & 0.0220 & 0.0397 \\
 &  DeCA  & 0.1967 & 0.2541 & 0.1218 & 0.1373 & 0.0590 & 0.1698 & 0.0460 & 0.0810 & 0.0214 & 0.0668 & 0.0156 & 0.0286 \\
 &  SGDL  & 0.1896 & 0.2565 & 0.1123 & 0.1304 & \underline{0.0748} & \underline{0.1981} & 0.0581 & \underline{0.0949} & 0.0290 & 0.0889 & 0.0209 & 0.0381 \\
 &  TCL  & \underline{0.1997} & 0.2601 & \underline{0.1232} & \underline{0.1396} & 0.0738 & 0.1746 & \underline{0.0600} & 0.0910 & \underline{0.0336} & \underline{0.1020} & \underline{0.0244} & \underline{0.0437} \\
 &  C2D  & 0.1893 & 0.2598 & 0.1176 & 0.1366 & 0.0727 & 0.1715 & 0.0594 & 0.0891 & 0.0312 & 0.0926 & 0.0223 & 0.0397 \\
\cmidrule(lr){2-14}
 &  \textbf{RGBT (Ours)}  & \textbf{0.2562} & \textbf{0.4177} & \textbf{0.1602} & \textbf{0.2002} & \textbf{0.0880} & \textbf{0.2039} & \textbf{0.0767} & \textbf{0.1115} & \textbf{0.0411} & \textbf{0.1275} & \textbf{0.0301} & \textbf{0.0543} \\
\midrule
\multirow{7}{*}{XSimGCL}  &  Normal  & 0.1071 & 0.1851 & 0.0649 & 0.0847 & 0.0231 & 0.0677 & 0.0149 & 0.0287 & 0.0584 & 0.1668 & 0.0435 & 0.0739 \\
 &  WBPR  & 0.1085 & \underline{0.1877} & 0.0665 & 0.0865 & 0.0253 & 0.0685 & 0.0168 & 0.0309 & 0.0547 & 0.1550 & 0.0410 & 0.0693 \\
 &  DeCA  & \underline{0.1244} & 0.1648 & \underline{0.0766} & \underline{0.0874} & \underline{0.0488} & \underline{0.1784} & \underline{0.0422} & \underline{0.0796} & 0.0403 & 0.1160 & 0.0306 & 0.0521 \\
 &  SGDL  & 0.1069 & 0.1867 & 0.0651 & 0.0853 & 0.0211 & 0.0641 & 0.0136 & 0.0275 & 0.0594 & 0.1672 & 0.0440 & 0.0742 \\
 &  TCL  & 0.0871 & 0.1769 & 0.0554 & 0.0781 & 0.0353 & 0.0985 & 0.0257 & 0.0454 & \underline{0.0605} & 0.1721 & \underline{0.0448} & 0.0760 \\
 &  C2D  & 0.1118 & 0.1844 & 0.0681 & 0.0864 & 0.0296 & 0.0848 & 0.0211 & 0.0382 & \textbf{0.0611} & \underline{0.1751} & \textbf{0.0454} & \textbf{0.0773} \\
\cmidrule(lr){2-14}
 &  \textbf{RGBT (Ours)}  & \textbf{0.2310} & \textbf{0.3118} & \textbf{0.1547} & \textbf{0.1766} & \textbf{0.0882} & \textbf{0.2146} & \textbf{0.0764} & \textbf{0.1150} & 0.0598 & \textbf{0.1761} & 0.0436 & \underline{0.0767} \\
\midrule
\multirow{7}{*}{CDAE}  &  Normal  & 0.2708 & 0.3656 & 0.1575 & 0.1833 & \underline{0.0761} & 0.1990 & 0.0626 & 0.0996 & \underline{0.0296} & \underline{0.0981} & \underline{0.0219} & \underline{0.0413} \\
 &  WBPR  & 0.2520 & 0.3616 & 0.1593 & 0.1905 & 0.0740 & 0.1906 & 0.0601 & 0.0960 & 0.0187 & 0.0619 & 0.0136 & 0.0261 \\
 &  DeCA  & 0.2354 & 0.3576 & 0.1561 & 0.1908 & 0.0438 & 0.1527 & 0.0368 & 0.0715 & 0.0207 & 0.0592 & 0.0153 & 0.0263 \\
 &  SGDL  & 0.2484 & 0.3735 & 0.1525 & 0.1860 & 0.0664 & 0.1951 & 0.0589 & 0.0971 & 0.0249 & 0.0809 & 0.0179 & 0.0339 \\
 &  TCL  & 0.1861 & \underline{0.4407} & 0.1093 & 0.1760 & 0.0677 & 0.2095 & 0.0541 & 0.0965 & 0.0268 & 0.0861 & 0.0195 & 0.0365 \\
 &  C2D  & \underline{0.2816} & \textbf{0.4895} & \underline{0.1750} & \textbf{0.2228} & 0.0682 & \underline{0.2106} & \underline{0.0628} & \underline{0.1049} & 0.0247 & 0.0826 & 0.0182 & 0.0348 \\
\cmidrule(lr){2-14}
 &  \textbf{RGBT (Ours)}  & \textbf{0.2886} & 0.3726 & \textbf{0.1761} & \underline{0.1986} & \textbf{0.0902} & \textbf{0.2178} & \textbf{0.0751} & \textbf{0.1130} & \textbf{0.0421} & \textbf{0.1347} & \textbf{0.0305} & \textbf{0.0567} \\
\bottomrule
\end{tabular*}
\end{table}

\FloatBarrier
\subsection{Matrix Estimation and Debiasing (RQ2)}

To answer RQ2, we strictly evaluate the estimation accuracy of the transition matrix using the synthetic flip noise settings described in Section~\ref{sec:noise_simulation}. We measure the $L_1$ norm discrepancy between the estimated transition matrix $\hat{T}$ and the ground-truth transition matrix $T$ on the MovieLens dataset.

\begin{table}[H]
\begin{minipage}[t]{0.48\linewidth}
\vspace{0pt}
\fontsize{8}{9.2}\selectfont
\renewcommand{\arraystretch}{1.0}
\centering
\caption{Matrix estimation error (mean $\pm$ standard deviation) over four backbones and four noise rates. Bold marks the lowest mean shown.}
\label{tab:l1_avg}
\setlength{\tabcolsep}{2pt}
\begin{tabular*}{\linewidth}{@{\extracolsep{\fill}}lcc@{}}
\toprule
Method & Pairflip & Symmetric \\
\midrule
CCR & 2.4474 $\pm$ 1.1558 & 2.2820 $\pm$ 1.1471 \\
DUAL\_T & 2.5983 $\pm$ 0.9354 & 2.2781 $\pm$ 0.9203 \\
T\_REVISION & 5.0203 $\pm$ 1.0717 & 4.6194 $\pm$ 1.2125 \\
VOLMINNET & 6.2523 $\pm$ 0.4488 & 5.4986 $\pm$ 1.1517 \\
RRFN & 2.2616 $\pm$ 1.0923 & 2.1965 $\pm$ 1.0764 \\
CONL & 3.4056 $\pm$ 0.4232 & 2.3971 $\pm$ 0.7384 \\
\cmidrule{1-3} 
RGBT (Ours) & \textbf{2.2240} $\boldsymbol{\pm}$ \textbf{0.9949} & \textbf{1.1050} $\boldsymbol{\pm}$ \textbf{0.6192} \\
\bottomrule
\end{tabular*}
\end{minipage}\hfill
\begin{minipage}[t]{0.48\linewidth}
\vspace{0pt}
\fontsize{7.2}{8.4}\selectfont
\setlength{\tabcolsep}{1.2pt}
\renewcommand{\arraystretch}{1.0}
\centering
\caption{Ablation study on ML-100k. Bold indicates the best performance, including ties.}
\label{tab:ablation_results}

    \begin{tabular*}{\linewidth}{@{\extracolsep{\fill}}llrrrr@{}}
    \toprule
    \multirow{2}{*}{Model} & \multirow{2}{*}{Variant} & \multicolumn{2}{c}{Recall} & \multicolumn{2}{c}{NDCG} \\
    \cmidrule(lr){3-4} \cmidrule(lr){5-6}
     & & @10 & @50 & @10 & @50 \\
    \midrule
    \multirow{4}{*}{NeuMF-end}
      &  Base  & 0.0811 & 0.2099 & 0.0698 & 0.1091 \\
      &  w/o BLTM  & 0.0824 & 0.2061 & 0.0710 & 0.1088 \\
      &  w/o GMM  & 0.0868 & \textbf{0.2155} & 0.0741 & 0.1125 \\
    \cmidrule(lr){2-6}
      &  \textbf{Full (Ours)}  & \textbf{0.0879} & 0.2151 & \textbf{0.0745} & \textbf{0.1130} \\
    \midrule
    \multirow{4}{*}{GMF}
      &  Base  & 0.0782 & 0.2046 & 0.0643 & 0.1041 \\
      &  w/o BLTM  & 0.0762 & 0.2053 & 0.0622 & 0.1026 \\
      &  w/o GMM  & 0.0806 & 0.2063 & \textbf{0.0671} & 0.1058 \\
    \cmidrule(lr){2-6}
      &  \textbf{Full (Ours)}  & \textbf{0.0836} & \textbf{0.2124} & 0.0664 & \textbf{0.1069} \\
    \bottomrule
    \end{tabular*}
\end{minipage}
\end{table}

\begin{figure}[H]
\begin{minipage}[t]{0.50\linewidth}
\vspace{0pt}
\centering
\includegraphics[width=\linewidth]{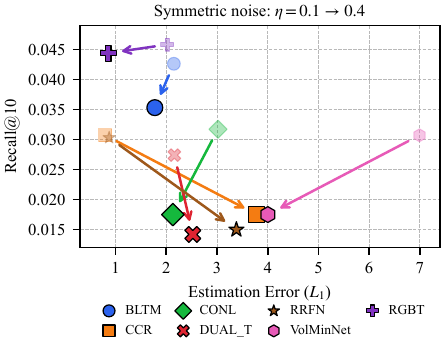}
\caption{Performance shift under symmetric noise from $\eta=0.1$ to $\eta=0.4$.}
\label{fig:noise_shift}
\end{minipage}\hfill
\begin{minipage}[t]{0.46\linewidth}
\vspace{0pt}
RGBT achieves the lowest average $L_1$ error among the methods in Table~\ref{tab:l1_avg} under both Pairflip noise (2.2240) and Symmetric noise (1.1050).

In Appendix~\ref{app:baseline_results}, Table~\ref{tab:main_noise_levels}, RGBT achieves the lowest mean error at Pairflip rates of 0.2 and 0.3, and at Symmetric rates of 0.2, 0.3, and 0.4. These results support effective estimation across several noise settings, but do not establish uniform superiority across all settings.

Figure~\ref{fig:noise_shift} shows that, under Symmetric noise, mean Recall@10 changes from 0.0458 to 0.0444 as $\eta$ increases from 0.1 to 0.4, while mean matrix error decreases from 2.0184 to 0.8554.
\end{minipage}
\end{figure}

Downstream recommendation also depends on the backbone and noise level. Under Pairflip noise, LightGCN obtains Recall@10 of 0.0568 at $\eta=0.1$ and 0.0556 at $\eta=0.4$, whereas NeuMF decreases from 0.0103 to 0.0029. Favorable matrix estimation results therefore do not imply uniform robustness of downstream recommendation across backbones.

\subsection{Parameter Sensitivity (RQ3)}
\label{sec:rq3}
\begin{figure}[H]
\begin{minipage}[t]{0.50\linewidth}
\vspace{0pt}
\centering
\includegraphics[width=\linewidth]{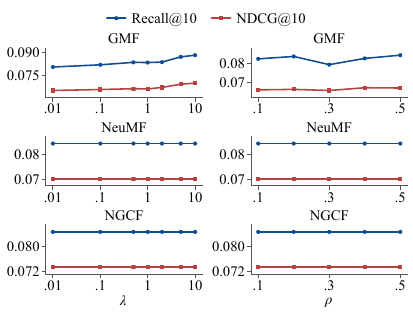}
\caption{Sensitivity to $\lambda$ (left, log scale) and $\rho$ (right) on ML-100k.}
\label{fig:sensitivity}
\end{minipage}\hfill
\begin{minipage}[t]{0.46\linewidth}
\vspace{0pt}
We investigate the calibrated classifier loss weight $\lambda$ and the distillation parameter $\rho$ on ML-100k (Figure~\ref{fig:sensitivity}). The confidence threshold is $(1+\rho)/2$.

For GMF, Recall@10 and NDCG@10 generally increase toward larger calibration weights within the tested range, with the highest reported values at $\lambda=10$.

The response to $\rho$ is nonmonotonic: GMF's Recall@10 is highest at $\rho=0.5$, whereas NDCG@10 is highest at $\rho=0.4$. NeuMF and NGCF show flat curves over both displayed ranges. Thus, the preferred setting depends on the backbone and metric.
\end{minipage}
\end{figure}

\subsection{Ablation Study (RQ4)}

We compare the full RGBT against Base, which uses standard classifier training, and two variants: (1)~\textit{w/o GMM}, which removes reliability weighting; and (2)~\textit{w/o BLTM}, which removes transition correction.

The full RGBT outperforms Base on all reported metrics for both backbones (Table~\ref{tab:ablation_results}).

\FloatBarrier
\endgroup

\section{Related Work}

\subsection{Denoising in Recommender Systems}
Early work on recommender systems relied on matrix factorization and graph models \cite{MF,WRMF,CDAE,LightGCN,KGAT,HOP-rec}. Recent efforts have focused on denoising, with approaches based on pruning \cite{ADT,RGCF}, GMMs \cite{JMPBP}, meta-learning \cite{SGDL,BOD,AutoDenoise}, and diffusion \cite{DiffRec,DDRM}. Other work has addressed exposure bias \cite{OME,WIPS,DCF}, contextual noise \cite{DSAT,CTR}, and negative sampling \cite{RNS,KGPolicy,EATNN}. Most denoising methods discard noisy examples, which can reduce data efficiency.

\subsection{Denoising in Statistical Methods}

Theoretical work includes PAC learning \cite{RCN}, consistent loss correction \cite{CCN}, Bayesian ranking \cite{BPR,WBPR,SBPR,IBPR}, and matrix factorization \cite{eALS}. Transition matrix methods have been explored via volume minimization \cite{VolMinNet}, part-dependent \cite{PTD} and dual-T \cite{Dual-T} approximations, instance-dependent models \cite{CSIDN,CoNL,IDNT,BIDN,STMN}, Bayesian correction \cite{NBPO}, transition revision \cite{T-Revision}, semi-supervised separation \cite{DivideMix}, bounded noise \cite{BILN}, causal learning \cite{CausalNL}, Bayes-optimal estimation \cite{BLTM,BLTM-V}, and cycle consistency \cite{CCR}. BLTM-based methods use all data but are often biased.

\subsection{Denoising in Machine Learning}
Machine learning has produced various noise-robust classification methods, from SVM \cite{SVM} and PU learning \cite{PU} to loss correction \cite{FC} and robust training \cite{fBGD,ITLM,REL,SELF,RSA,AT,PES,NGCF}. Benchmarks \cite{CIFAR-N} and causal studies \cite{SSL} have also been developed. Theoretical guarantees are limited.

\section{Conclusion}

We propose RGBT, a framework that uses GMM-based reliability weights to calibrate the Bayes-label transition matrix for recommendation with instance-dependent noise. The GMM assigns each instance a reliability score, and these scores adjust the transition matrix to reduce bias while using all available data. We show theoretically that RGBT uses all data, gives consistent estimates, and reduces variance compared to existing methods, extending beyond GMM-based calibration to any method with deterministic labels and non-negative weights. Experiments on real and synthetic data confirm that RGBT handles noise better than sample-selection baselines and calibrates the transition matrix more accurately than existing approaches. This work suggests that combining reliability weighting with transition matrix estimation is a promising direction for robust recommendation. Future work could extend this framework by incorporating causal modeling of noise feedback.

\clearpage
\section*{LLM Usage Disclosure}

We did not use any large language model (LLM) for research tasks such as ideation, experimental design, mathematical reasoning, or code implementation. Generative AI tools were used only for minor language polishing and grammar correction during the paper writing process. All technical content, claims, and artifacts are the sole work of the authors, who take full responsibility for the final paper.

\bibliography{iclr2027_conference}

\begin{thebibliography}{72}
\providecommand{\natexlab}[1]{#1}
\providecommand{\url}[1]{\texttt{#1}}
\expandafter\ifx\csname urlstyle\endcsname\relax
  \providecommand{\doi}[1]{doi: #1}\else
  \providecommand{\doi}{doi: \begingroup \urlstyle{rm}\Url}\fi

\bibitem[Angluin \& Laird(1988)Angluin and Laird]{RCN}
Dana Angluin and Philip Laird.
\newblock Learning from noisy examples.
\newblock \emph{Machine learning}, 2\penalty0 (4):\penalty0 343--370, 1988.

\bibitem[Bai et~al.(2024)Bai, Hou, Wu, Yang, Zhang, Hong, and
  Wang]{Recommender_noise_1}
Haoyue Bai, Min Hou, Le~Wu, Yonghui Yang, Kun Zhang, Richang Hong, and Meng
  Wang.
\newblock Unified representation learning for discrete attribute enhanced
  completely cold-start recommendation.
\newblock \emph{IEEE Transactions on Big Data}, 2024.

\bibitem[Bai et~al.(2021)Bai, Yang, Han, Yang, Li, Mao, Niu, and Liu]{PES}
Yingbin Bai, Erkun Yang, Bo~Han, Yanhua Yang, Jiatong Li, Yinian Mao, Gang Niu,
  and Tongliang Liu.
\newblock Understanding and improving early stopping for learning with noisy
  labels.
\newblock \emph{Advances in Neural Information Processing Systems},
  34:\penalty0 24392--24403, 2021.

\bibitem[Bai et~al.(2022)Bai, Yang, Wang, Du, Han, Deng, Wang, and Liu]{RSA}
Yingbin Bai, Erkun Yang, Zhaoqing Wang, Yuxuan Du, Bo~Han, Cheng Deng, Dadong
  Wang, and Tongliang Liu.
\newblock Rsa: reducing semantic shift from aggressive augmentations for
  self-supervised learning.
\newblock \emph{Advances in Neural Information Processing Systems},
  35:\penalty0 21128--21141, 2022.

\bibitem[Berthon et~al.(2021)Berthon, Han, Niu, Liu, and Sugiyama]{CSIDN}
Antonin Berthon, Bo~Han, Gang Niu, Tongliang Liu, and Masashi Sugiyama.
\newblock Confidence scores make instance-dependent label-noise learning
  possible.
\newblock In \emph{International conference on machine learning}, pp.\
  825--836. PMLR, 2021.

\bibitem[Cai et~al.(2024)Cai, Hou, Chen, Wu, Bai, Li, and
  Wang]{Recommender_noise_2}
Miaomiao Cai, Min Hou, Lei Chen, Le~Wu, Haoyue Bai, Yong Li, and Meng Wang.
\newblock Mitigating recommendation biases via group-alignment and
  global-uniformity in representation learning.
\newblock \emph{ACM Transactions on Intelligent Systems and Technology},
  15\penalty0 (5):\penalty0 1--27, 2024.

\bibitem[Cao et~al.(2024)Cao, Xu, Yang, Ma, and Huang]{WIPS}
Tianwei Cao, Qianqian Xu, Zhiyong Yang, Zhanyu Ma, and Qingming Huang.
\newblock Practically unbiased pairwise loss for recommendation with implicit
  feedback.
\newblock \emph{IEEE Transactions on Pattern Analysis and Machine
  Intelligence}, 2024.

\bibitem[Chen et~al.(2019)Chen, Zhang, Wang, Ma, Li, Liu, and Ma]{EATNN}
Chong Chen, Min Zhang, Chenyang Wang, Weizhi Ma, Minming Li, Yiqun Liu, and
  Shaoping Ma.
\newblock An efficient adaptive transfer neural network for social-aware
  recommendation.
\newblock In \emph{Proceedings of the 42nd international ACM SIGIR conference
  on research and development in information retrieval}, pp.\  225--234, 2019.

\bibitem[Cheng et~al.(2022{\natexlab{a}})Cheng, Liu, Ning, Wang, Han, Niu, Gao,
  and Sugiyama]{MEIDTM}
De~Cheng, Tongliang Liu, Yixiong Ning, Nannan Wang, Bo~Han, Gang Niu, Xinbo
  Gao, and Masashi Sugiyama.
\newblock Instance-dependent label-noise learning with manifold-regularized
  transition matrix estimation.
\newblock In \emph{Proceedings of the IEEE/CVF Conference on Computer Vision
  and Pattern Recognition}, pp.\  16630--16639, 2022{\natexlab{a}}.

\bibitem[Cheng et~al.(2022{\natexlab{b}})Cheng, Ning, Wang, Gao, Yang, Du, Han,
  and Liu]{CCR}
De~Cheng, Yixiong Ning, Nannan Wang, Xinbo Gao, Heng Yang, Yuxuan Du, Bo~Han,
  and Tongliang Liu.
\newblock Class-dependent label-noise learning with cycle-consistency
  regularization.
\newblock \emph{Advances in Neural Information Processing Systems},
  35:\penalty0 11104--11116, 2022{\natexlab{b}}.

\bibitem[Cheng et~al.(2020)Cheng, Liu, Ramamohanarao, and Tao]{BILN}
Jiacheng Cheng, Tongliang Liu, Kotagiri Ramamohanarao, and Dacheng Tao.
\newblock Learning with bounded instance and label-dependent label noise.
\newblock In \emph{International conference on machine learning}, pp.\
  1789--1799. PMLR, 2020.

\bibitem[Ding et~al.(2018)Ding, Feng, He, Yu, Li, and Jin]{IBPR}
Jingtao Ding, Fuli Feng, Xiangnan He, Guanghui Yu, Yong Li, and Depeng Jin.
\newblock An improved sampler for bayesian personalized ranking by leveraging
  view data.
\newblock In \emph{Companion proceedings of the the web conference 2018}, pp.\
  13--14, 2018.

\bibitem[Ding et~al.(2019{\natexlab{a}})Ding, Quan, He, Li, and Jin]{RNS}
Jingtao Ding, Yuhan Quan, Xiangnan He, Yong Li, and Depeng Jin.
\newblock Reinforced negative sampling for recommendation with exposure data.
\newblock In \emph{IJCAI}, pp.\  2230--2236. Macao, 2019{\natexlab{a}}.

\bibitem[Ding et~al.(2019{\natexlab{b}})Ding, Yu, He, Feng, Li, and Jin]{SBPR}
Jingtao Ding, Guanghui Yu, Xiangnan He, Fuli Feng, Yong Li, and Depeng Jin.
\newblock Sampler design for bayesian personalized ranking by leveraging view
  data.
\newblock \emph{IEEE transactions on knowledge and data engineering},
  33\penalty0 (2):\penalty0 667--681, 2019{\natexlab{b}}.

\bibitem[Du~Plessis et~al.(2014)Du~Plessis, Niu, and Sugiyama]{PU}
Marthinus~C Du~Plessis, Gang Niu, and Masashi Sugiyama.
\newblock Analysis of learning from positive and unlabeled data.
\newblock \emph{Advances in neural information processing systems}, 27, 2014.

\bibitem[Gantner et~al.(2012)Gantner, Drumond, Freudenthaler, and
  Schmidt-Thieme]{WBPR}
Zeno Gantner, Lucas Drumond, Christoph Freudenthaler, and Lars Schmidt-Thieme.
\newblock Personalized ranking for non-uniformly sampled items.
\newblock In \emph{Proceedings of KDD cup 2011}, pp.\  231--247. PMLR, 2012.

\bibitem[Gao et~al.(2022)Gao, Du, Hu, Chen, Zhu, Fang, and Zheng]{SGDL}
Yunjun Gao, Yuntao Du, Yujia Hu, Lu~Chen, Xinjun Zhu, Ziquan Fang, and Baihua
  Zheng.
\newblock Self-guided learning to denoise for robust recommendation.
\newblock In \emph{Proceedings of the 45th international ACM SIGIR conference
  on research and development in information retrieval}, pp.\  1412--1422,
  2022.

\bibitem[Ge et~al.(2023)Ge, Rahmani, Irissappane, Sepulveda, Caverlee, and
  Wang]{AutoDenoise}
Yingqiang Ge, Mostafa Rahmani, Athirai~A Irissappane, Jose Sepulveda, James
  Caverlee, and Fei Wang.
\newblock Automated data denoising for recommendation.
\newblock \emph{CoRR}, 2023.

\bibitem[He \& Chua(2017)He and Chua]{GMF+NeuMF}
Xiangnan He and Tat-Seng Chua.
\newblock Neural factorization machines for sparse predictive analytics.
\newblock In \emph{Proceedings of the 40th International ACM SIGIR conference
  on Research and Development in Information Retrieval}, pp.\  355--364, 2017.

\bibitem[He et~al.(2016)He, Zhang, Kan, and Chua]{eALS}
Xiangnan He, Hanwang Zhang, Min-Yen Kan, and Tat-Seng Chua.
\newblock Fast matrix factorization for online recommendation with implicit
  feedback.
\newblock In \emph{Proceedings of the 39th International ACM SIGIR conference
  on Research and Development in Information Retrieval}, pp.\  549--558, 2016.

\bibitem[He et~al.(2020)He, Deng, Wang, Li, Zhang, and Wang]{LightGCN}
Xiangnan He, Kuan Deng, Xiang Wang, Yan Li, Yongdong Zhang, and Meng Wang.
\newblock Lightgcn: Simplifying and powering graph convolution network for
  recommendation.
\newblock In \emph{Proceedings of the 43rd International ACM SIGIR conference
  on research and development in Information Retrieval}, pp.\  639--648, 2020.

\bibitem[He et~al.(2024)He, Wang, Yang, Sun, Wu, Bai, Gong, Hong, and
  Zhang]{DCF}
Zhuangzhuang He, Yifan Wang, Yonghui Yang, Peijie Sun, Le~Wu, Haoyue Bai, Jinqi
  Gong, Richang Hong, and Min Zhang.
\newblock Double correction framework for denoising recommendation.
\newblock In \emph{Proceedings of the 30th ACM SIGKDD Conference on Knowledge
  Discovery and Data Mining}, pp.\  1062--1072, 2024.

\bibitem[Hu et~al.(2008)Hu, Koren, and Volinsky]{WRMF}
Yifan Hu, Yehuda Koren, and Chris Volinsky.
\newblock Collaborative filtering for implicit feedback datasets.
\newblock In \emph{2008 Eighth IEEE international conference on data mining},
  pp.\  263--272. Ieee, 2008.

\bibitem[Kim et~al.(2014)Kim, Hassan, White, and Zitouni]{DSAT}
Youngho Kim, Ahmed Hassan, Ryen~W White, and Imed Zitouni.
\newblock Modeling dwell time to predict click-level satisfaction.
\newblock In \emph{Proceedings of the 7th ACM international conference on Web
  search and data mining}, pp.\  193--202, 2014.

\bibitem[Koren et~al.(2009)Koren, Bell, and Volinsky]{MF}
Yehuda Koren, Robert Bell, and Chris Volinsky.
\newblock Matrix factorization techniques for recommender systems.
\newblock \emph{Computer}, 42\penalty0 (8):\penalty0 30--37, 2009.

\bibitem[Krauth et~al.(2025)Krauth, Wang, and Jordan]{Recommder_noise_6}
Karl Krauth, Yixin Wang, and Michael Jordan.
\newblock Breaking feedback loops in recommender systems with causal inference.
\newblock \emph{ACM Transactions on Recommender Systems}, 4\penalty0
  (1):\penalty0 1--20, 2025.

\bibitem[Li et~al.(2024{\natexlab{a}})Li, Zheng, Wang, Wang, Feng, and
  Zhou]{OME}
Haoxuan Li, Chunyuan Zheng, Wenjie Wang, Hao Wang, Fuli Feng, and Xiao-Hua
  Zhou.
\newblock Debiased recommendation with noisy feedback.
\newblock In \emph{Proceedings of the 30th ACM SIGKDD Conference on Knowledge
  Discovery and Data Mining}, pp.\  1576--1586, 2024{\natexlab{a}}.

\bibitem[Li et~al.(2020)Li, Socher, and Hoi]{DivideMix}
Junnan Li, Richard Socher, and Steven~CH Hoi.
\newblock Dividemix: Learning with noisy labels as semi-supervised learning.
\newblock In \emph{International Conference on Learning Representations}, 2020.

\bibitem[Li et~al.(2021)Li, Liu, Han, Niu, and Sugiyama]{VolMinNet}
Xuefeng Li, Tongliang Liu, Bo~Han, Gang Niu, and Masashi Sugiyama.
\newblock Provably end-to-end label-noise learning without anchor points.
\newblock In \emph{International conference on machine learning}, pp.\
  6403--6413. PMLR, 2021.

\bibitem[Li et~al.(2024{\natexlab{b}})Li, Liu, Satapathy, Wang, and
  Cambria]{Recommender_survey_3}
Yang Li, Kangbo Liu, Ranjan Satapathy, Suhang Wang, and Erik Cambria.
\newblock Recent developments in recommender systems: A survey.
\newblock \emph{IEEE Computational Intelligence Magazine}, 19\penalty0
  (2):\penalty0 78--95, 2024{\natexlab{b}}.

\bibitem[Lin et~al.(2025)Lin, Yao, Wang, Shen, Yu, Han, and Liu]{CoNL}
Yexiong Lin, Yu~Yao, Zhaoqing Wang, Xu~Shen, Jun Yu, Bo~Han, and Tongliang Liu.
\newblock Improving the instance-dependent transition matrix estimation by
  exploiting self-supervised learning.
\newblock \emph{IEEE Transactions on Pattern Analysis and Machine
  Intelligence}, 2025.

\bibitem[Liu et~al.(2024{\natexlab{a}})Liu, Wang, Zhang, Deng, Chen, and
  Zhang]{Recommder_noise_4}
Hanyang Liu, Yong Wang, Zhiqiang Zhang, Jiangzhou Deng, Chao Chen, and Leo~Yu
  Zhang.
\newblock Matrix factorization recommender based on adaptive gaussian
  differential privacy for implicit feedback.
\newblock \emph{Information Processing \& Management}, 61\penalty0
  (4):\penalty0 103720, 2024{\natexlab{a}}.

\bibitem[Liu et~al.(2024{\natexlab{b}})Liu, Hu, Xiao, Zhao, Gao, Wang, Li, and
  Tang]{Recommender_survey_1}
Qidong Liu, Jiaxi Hu, Yutian Xiao, Xiangyu Zhao, Jingtong Gao, Wanyu Wang, Qing
  Li, and Jiliang Tang.
\newblock Multimodal recommender systems: A survey.
\newblock \emph{ACM Computing Surveys}, 57\penalty0 (2):\penalty0 1--17,
  2024{\natexlab{b}}.

\bibitem[Liu \& Tao(2015)Liu and Tao]{CCN}
Tongliang Liu and Dacheng Tao.
\newblock Classification with noisy labels by importance reweighting.
\newblock \emph{IEEE Transactions on pattern analysis and machine
  intelligence}, 38\penalty0 (3):\penalty0 447--461, 2015.

\bibitem[Lu et~al.(2018)Lu, Zhang, and Ma]{CTR}
Hongyu Lu, Min Zhang, and Shaoping Ma.
\newblock Between clicks and satisfaction: Study on multi-phase user
  preferences and satisfaction for online news reading.
\newblock In \emph{The 41st international acm sigir conference on research \&
  development in information retrieval}, pp.\  435--444, 2018.

\bibitem[Menon et~al.(2018)Menon, Van~Rooyen, and Natarajan]{BIDN}
Aditya~Krishna Menon, Brendan Van~Rooyen, and Nagarajan Natarajan.
\newblock Learning from binary labels with instance-dependent noise.
\newblock \emph{Machine Learning}, 107\penalty0 (8):\penalty0 1561--1595, 2018.

\bibitem[Natarajan et~al.(2013)Natarajan, Dhillon, Ravikumar, and Tewari]{SVM}
Nagarajan Natarajan, Inderjit~S Dhillon, Pradeep~K Ravikumar, and Ambuj Tewari.
\newblock Learning with noisy labels.
\newblock \emph{Advances in neural information processing systems}, 26, 2013.

\bibitem[Nguyen et~al.(2020)Nguyen, Mummadi, Ngo, Nguyen, Beggel, and
  Brox]{SELF}
Duc~Tam Nguyen, Chaithanya~Kumar Mummadi, Thi Phuong~Nhung Ngo, Thi Hoai~Phuong
  Nguyen, Laura Beggel, and Thomas Brox.
\newblock Self: Learning to filter noisy labels with self-ensembling.
\newblock In \emph{International Conference on Learning Representations}, 2020.

\bibitem[Patrini et~al.(2017)Patrini, Rozza, Krishna~Menon, Nock, and Qu]{FC}
Giorgio Patrini, Alessandro Rozza, Aditya Krishna~Menon, Richard Nock, and
  Lizhen Qu.
\newblock Making deep neural networks robust to label noise: A loss correction
  approach.
\newblock In \emph{Proceedings of the IEEE conference on computer vision and
  pattern recognition}, pp.\  1944--1952, 2017.

\bibitem[Rendle et~al.(2012)Rendle, Freudenthaler, Gantner, and
  Schmidt-Thieme]{BPR}
Steffen Rendle, Christoph Freudenthaler, Zeno Gantner, and Lars Schmidt-Thieme.
\newblock Bpr: Bayesian personalized ranking from implicit feedback.
\newblock \emph{arXiv preprint arXiv:1205.2618}, 2012.

\bibitem[Shen \& Sanghavi(2019)Shen and Sanghavi]{ITLM}
Yanyao Shen and Sujay Sanghavi.
\newblock Learning with bad training data via iterative trimmed loss
  minimization.
\newblock In \emph{International conference on machine learning}, pp.\
  5739--5748. PMLR, 2019.

\bibitem[Sugiyama et~al.(2025)Sugiyama, Yoshida, and Muneyasu]{JMPBP}
Takuto Sugiyama, Soh Yoshida, and Mitsuji Muneyasu.
\newblock Joint modeling of prediction and behavioral patterns for reliable
  recommendation with implicit feedback.
\newblock \emph{IEEE Access}, 2025.

\bibitem[Sun et~al.(2024)Sun, Wang, Zhang, Wu, Fang, Zhu, Fang, and
  Wang]{Recommder_noise_3}
Peijie Sun, Yifan Wang, Min Zhang, Chuhan Wu, Yan Fang, Hong Zhu, Yuan Fang,
  and Meng Wang.
\newblock Collaborative-enhanced prediction of spending on newly downloaded
  mobile games under consumption uncertainty.
\newblock In \emph{Companion Proceedings of the ACM Web Conference 2024}, pp.\
  10--19, 2024.

\bibitem[Tian et~al.(2022)Tian, Xie, Li, Yang, and Zhao]{RGCF}
Changxin Tian, Yuexiang Xie, Yaliang Li, Nan Yang, and Wayne~Xin Zhao.
\newblock Learning to denoise unreliable interactions for graph collaborative
  filtering.
\newblock In \emph{Proceedings of the 45th international ACM SIGIR conference
  on research and development in information retrieval}, pp.\  122--132, 2022.

\bibitem[Wang et~al.(2021)Wang, Feng, He, Nie, and Chua]{ADT}
Wenjie Wang, Fuli Feng, Xiangnan He, Liqiang Nie, and Tat-Seng Chua.
\newblock Denoising implicit feedback for recommendation.
\newblock In \emph{Proceedings of the 14th ACM international conference on web
  search and data mining}, pp.\  373--381, 2021.

\bibitem[Wang et~al.(2023{\natexlab{a}})Wang, Xu, Feng, Lin, He, and
  Chua]{DiffRec}
Wenjie Wang, Yiyan Xu, Fuli Feng, Xinyu Lin, Xiangnan He, and Tat-Seng Chua.
\newblock Diffusion recommender model.
\newblock In \emph{Proceedings of the 46th international ACM SIGIR conference
  on research and development in information retrieval}, pp.\  832--841,
  2023{\natexlab{a}}.

\bibitem[Wang et~al.(2023{\natexlab{b}})Wang, Zhang, Li, Wu, Feng, and
  He]{Recommender_survey_2}
Wenjie Wang, Yang Zhang, Haoxuan Li, Peng Wu, Fuli Feng, and Xiangnan He.
\newblock Causal recommendation: Progresses and future directions.
\newblock In \emph{Proceedings of the 46th international ACM SIGIR conference
  on research and development in information retrieval}, pp.\  3432--3435,
  2023{\natexlab{b}}.

\bibitem[Wang et~al.(2019{\natexlab{a}})Wang, He, Cao, Liu, and Chua]{KGAT}
Xiang Wang, Xiangnan He, Yixin Cao, Meng Liu, and Tat-Seng Chua.
\newblock Kgat: Knowledge graph attention network for recommendation.
\newblock In \emph{Proceedings of the 25th ACM SIGKDD international conference
  on knowledge discovery \& data mining}, pp.\  950--958, 2019{\natexlab{a}}.

\bibitem[Wang et~al.(2019{\natexlab{b}})Wang, He, Wang, Feng, and Chua]{NGCF}
Xiang Wang, Xiangnan He, Meng Wang, Fuli Feng, and Tat-Seng Chua.
\newblock Neural graph collaborative filtering.
\newblock In \emph{Proceedings of the 42nd international ACM SIGIR conference
  on Research and development in Information Retrieval}, pp.\  165--174,
  2019{\natexlab{b}}.

\bibitem[Wang et~al.(2020)Wang, Xu, He, Cao, Wang, and Chua]{KGPolicy}
Xiang Wang, Yaokun Xu, Xiangnan He, Yixin Cao, Meng Wang, and Tat-Seng Chua.
\newblock Reinforced negative sampling over knowledge graph for recommendation.
\newblock In \emph{Proceedings of the web conference 2020}, pp.\  99--109,
  2020.

\bibitem[Wang et~al.(2024)Wang, Sun, Ma, Zhang, Zhang, Jiang, and
  Ma]{Recommder_noise_5}
Yifan Wang, Peijie Sun, Weizhi Ma, Min Zhang, Yuan Zhang, Peng Jiang, and
  Shaoping Ma.
\newblock Intersectional two-sided fairness in recommendation.
\newblock In \emph{Proceedings of the ACM Web Conference 2024}, pp.\
  3609--3620, 2024.

\bibitem[Wang et~al.(2022)Wang, Xin, Meng, Jose, Feng, and He]{DeCA}
Yu~Wang, Xin Xin, Zaiqiao Meng, Joemon~M Jose, Fuli Feng, and Xiangnan He.
\newblock Learning robust recommenders through cross-model agreement.
\newblock In \emph{Proceedings of the ACM web conference 2022}, pp.\
  2015--2025, 2022.

\bibitem[Wang et~al.(2025)Wang, Pang, Qin, Wei, and Zhao]{IDNT}
Yuan Wang, Huaxin Pang, Ying Qin, Shikui Wei, and Yao Zhao.
\newblock Adaptive estimation of instance-dependent noise transition matrix for
  learning with instance-dependent label noise.
\newblock \emph{Neural Networks}, 188:\penalty0 107464, 2025.

\bibitem[Wang et~al.(2023{\natexlab{c}})Wang, Gao, Li, Yu, Guo, and Yin]{BOD}
Zongwei Wang, Min Gao, Wentao Li, Junliang Yu, Linxin Guo, and Hongzhi Yin.
\newblock Efficient bi-level optimization for recommendation denoising.
\newblock In \emph{Proceedings of the 29th ACM SIGKDD conference on knowledge
  discovery and data mining}, pp.\  2502--2511, 2023{\natexlab{c}}.

\bibitem[Wei et~al.(2022)Wei, Zhu, Cheng, Liu, Niu, and Liu]{CIFAR-N}
Jiaheng Wei, Zhaowei Zhu, Hao Cheng, Tongliang Liu, Gang Niu, and Yang Liu.
\newblock Learning with noisy labels revisited: A study using real-world human
  annotations.
\newblock In \emph{International Conference on Learning Representations}, 2022.

\bibitem[Wu et~al.(2016)Wu, DuBois, Zheng, and Ester]{CDAE}
Yao Wu, Christopher DuBois, Alice~X Zheng, and Martin Ester.
\newblock Collaborative denoising auto-encoders for top-n recommender systems.
\newblock In \emph{Proceedings of the ninth ACM international conference on web
  search and data mining}, pp.\  153--162, 2016.

\bibitem[Xia et~al.(2019)Xia, Liu, Wang, Han, Gong, Niu, and
  Sugiyama]{T-Revision}
Xiaobo Xia, Tongliang Liu, Nannan Wang, Bo~Han, Chen Gong, Gang Niu, and
  Masashi Sugiyama.
\newblock Are anchor points really indispensable in label-noise learning?
\newblock \emph{Advances in neural information processing systems}, 32, 2019.

\bibitem[Xia et~al.(2020{\natexlab{a}})Xia, Liu, Han, Gong, Wang, Ge, and
  Chang]{REL}
Xiaobo Xia, Tongliang Liu, Bo~Han, Chen Gong, Nannan Wang, Zongyuan Ge, and
  Yi~Chang.
\newblock Robust early-learning: Hindering the memorization of noisy labels.
\newblock In \emph{International conference on learning representations},
  2020{\natexlab{a}}.

\bibitem[Xia et~al.(2020{\natexlab{b}})Xia, Liu, Han, Wang, Gong, Liu, Niu,
  Tao, and Sugiyama]{PTD}
Xiaobo Xia, Tongliang Liu, Bo~Han, Nannan Wang, Mingming Gong, Haifeng Liu,
  Gang Niu, Dacheng Tao, and Masashi Sugiyama.
\newblock Part-dependent label noise: Towards instance-dependent label noise.
\newblock \emph{Advances in neural information processing systems},
  33:\penalty0 7597--7610, 2020{\natexlab{b}}.

\bibitem[Yang et~al.(2018)Yang, Chen, Wang, and Tsai]{HOP-rec}
Jheng-Hong Yang, Chih-Ming Chen, Chuan-Ju Wang, and Ming-Feng Tsai.
\newblock Hop-rec: high-order proximity for implicit recommendation.
\newblock In \emph{Proceedings of the 12th ACM conference on recommender
  systems}, pp.\  140--144, 2018.

\bibitem[Yang et~al.(2022)Yang, Yang, Han, Liu, Xu, Niu, and Liu]{BLTM}
Shuo Yang, Erkun Yang, Bo~Han, Yang Liu, Min Xu, Gang Niu, and Tongliang Liu.
\newblock Estimating instance-dependent bayes-label transition matrix using a
  deep neural network.
\newblock In \emph{International Conference on Machine Learning}, pp.\
  25302--25312. PMLR, 2022.

\bibitem[Yang et~al.(2023)Yang, Wu, Yang, Han, Liu, Xu, Niu, and Liu]{BLTM-V}
Shuo Yang, Songhua Wu, Erkun Yang, Bo~Han, Yang Liu, Min Xu, Gang Niu, and
  Tongliang Liu.
\newblock A parametrical model for instance-dependent label noise.
\newblock \emph{IEEE Transactions on Pattern Analysis and Machine
  Intelligence}, 45\penalty0 (12):\penalty0 14055--14068, 2023.

\bibitem[Yao et~al.(2020)Yao, Liu, Han, Gong, Deng, Niu, and Sugiyama]{Dual-T}
Yu~Yao, Tongliang Liu, Bo~Han, Mingming Gong, Jiankang Deng, Gang Niu, and
  Masashi Sugiyama.
\newblock Dual t: Reducing estimation error for transition matrix in
  label-noise learning.
\newblock \emph{Advances in neural information processing systems},
  33:\penalty0 7260--7271, 2020.

\bibitem[Yao et~al.(2021)Yao, Liu, Gong, Han, Niu, and Zhang]{CausalNL}
Yu~Yao, Tongliang Liu, Mingming Gong, Bo~Han, Gang Niu, and Kun Zhang.
\newblock Instance-dependent label-noise learning under a structural causal
  model.
\newblock \emph{Advances in Neural Information Processing Systems},
  34:\penalty0 4409--4420, 2021.

\bibitem[Yao et~al.(2023)Yao, Gong, Du, Yu, Han, Zhang, and Liu]{SSL}
Yu~Yao, Mingming Gong, Yuxuan Du, Jun Yu, Bo~Han, Kun Zhang, and Tongliang Liu.
\newblock Which is better for learning with noisy labels: the semi-supervised
  method or modeling label noise?
\newblock In \emph{International conference on machine learning}, pp.\
  39660--39673. PMLR, 2023.

\bibitem[Ye \& Lu(2024)Ye and Lu]{RRFN}
Shanshan Ye and Jie Lu.
\newblock Robust recommender systems with rating flip noise.
\newblock \emph{ACM Transactions on Intelligent Systems and Technology},
  16\penalty0 (1):\penalty0 1--19, 2024.

\bibitem[Yin et~al.(2025)Yin, Qu, Chen, Yuan, Zheng, Long, Xia, Shi, and
  Zhang]{Recommender_survey_4}
Hongzhi Yin, Liang Qu, Tong Chen, Wei Yuan, Ruiqi Zheng, Jing Long, Xin Xia,
  Yuhui Shi, and Chengqi Zhang.
\newblock On-device recommender systems: A comprehensive survey.
\newblock \emph{Data Science and Engineering}, pp.\  1--30, 2025.

\bibitem[Yu \& Qin(2020)Yu and Qin]{NBPO}
Wenhui Yu and Zheng Qin.
\newblock Sampler design for implicit feedback data by noisy-label robust
  learning.
\newblock In \emph{Proceedings of the 43rd international ACM SIGIR conference
  on research and development in information retrieval}, pp.\  861--870, 2020.

\bibitem[Yuan et~al.(2018)Yuan, Xin, He, Guo, Zhang, Tat-Seng, and Jose]{fBGD}
Fajie Yuan, Xin Xin, Xiangnan He, Guibing Guo, Weinan Zhang, Chua Tat-Seng, and
  Joemon~M Jose.
\newblock fbgd: Learning embeddings from positive unlabeled data with bgd.
\newblock 2018.

\bibitem[Zhang et~al.(2024)Zhang, Cao, Yang, Si, Sun, Xu, and Sun]{STMN}
Ruiheng Zhang, Zhe Cao, Shuo Yang, Lingyu Si, Haoyang Sun, Lixin Xu, and Fuchun
  Sun.
\newblock Cognition-driven structural prior for instance-dependent label
  transition matrix estimation.
\newblock \emph{IEEE Transactions on Neural Networks and Learning Systems},
  2024.

\bibitem[Zhao et~al.(2024)Zhao, Wenjie, Xu, Sun, Feng, and Chua]{DDRM}
Jujia Zhao, Wang Wenjie, Yiyan Xu, Teng Sun, Fuli Feng, and Tat-Seng Chua.
\newblock Denoising diffusion recommender model.
\newblock In \emph{Proceedings of the 47th International ACM SIGIR Conference
  on Research and Development in Information Retrieval}, pp.\  1370--1379,
  2024.

\bibitem[Zhu et~al.(2021)Zhu, Zhang, Han, Liu, Niu, Yang, Kankanhalli, and
  Sugiyama]{AT}
Jianing Zhu, Jingfeng Zhang, Bo~Han, Tongliang Liu, Gang Niu, Hongxia Yang,
  Mohan Kankanhalli, and Masashi Sugiyama.
\newblock Understanding the interaction of adversarial training with noisy
  labels.
\newblock \emph{arXiv preprint arXiv:2102.03482}, 2021.

\end{thebibliography}
\bibliographystyle{iclr2027_conference}

\appendix
\section{Theoretical Derivations}
\label{app:theory}

This appendix contains the derivations and proofs for the results in
Section~\ref{sec:theory}. We first state the identifiability condition
under which the transition matrix can be recovered at all. We then
derive the CLTM and BLTM estimators, prove
Theorem~\ref{thm:variance_reduction}, state the finite-sample bound,
and finally give a general condition under which the same
variance-reduction argument extends beyond GMM weighting.

\subsection{Manifold-Regularized Transition Matrix}
\label{app:manifold}

The variance-reduction result of Theorem~\ref{thm:variance_reduction}
compares two estimators under the assumption that $T^{*}(\mathbf{x})$
can be estimated stably. This assumption is not automatic under
instance-dependent noise, and we make it explicit here. The results
below state when the estimation and correction steps are well posed,
and serve as a precondition for the main text.

Under instance-dependent noise,
\begin{equation}
\widetilde{\eta}(\mathbf{x})
=
T^{*}(\mathbf{x})^{\top} \eta(\mathbf{x}),
\label{eq:idn_relation}
\end{equation}
where $\eta(\mathbf{x})$ is the clean posterior and
$T^{*}(\mathbf{x})$ is the instance-dependent transition matrix.

With a single noisy label per instance, $T^{*}(\mathbf{x})$ is not
identifiable from the data alone. We therefore adopt the
manifold-regularization assumption: the transition matrix is Lipschitz
continuous in the feature space, i.e., there exists $L>0$ such that
\begin{equation}
\|T^{*}(\mathbf{x}_i) - T^{*}(\mathbf{x}_j)\|
\le L \|\mathbf{x}_i - \mathbf{x}_j\|.
\label{eq:lipschitz}
\end{equation}

\begin{assumption}[Manifold-regularized transition matrix]
\label{ass:manifold}
The transition matrix $T^{*}(\mathbf{x})$ is Lipschitz continuous in
$\mathbf{x}$ with constant $L$.
\end{assumption}

Under Assumption~\ref{ass:manifold}, nearby instances have similar
transition matrices, which reduces the effective degrees of freedom of
$T^{*}(\mathbf{x})$ and makes the estimation and correction steps below
well posed. This assumption is natural in recommendation: similar
users and similar items exhibit similar noise patterns, which is
consistent with the reliability features used by the GMM in
Section~\ref{sec:theory}.

Throughout the rest of this appendix we assume that
Assumption~\ref{ass:manifold} holds.

\subsection{CLTM and BLTM Estimators}

Our derivation proceeds in two steps. The calibrated loss
\(\mathcal{L}_{\text{calibrated}}\) first drives the classifier toward the
clean posterior \(\eta(\mathbf{x})\). The classification loss
\(\mathcal{L}_{\text{class}}\) then estimates the transition matrix so that
\(\widehat{T}^* f\) matches the observed noisy posterior
\(\widetilde{\eta}\). Both steps are needed: without the first, \(f\) would
remain at the noisy posterior; without the second, the transition matrix
would not be identified.

\subsubsection{Calibrating the Classifier}

We first extract high-confidence distilled labels \(Y^*\) by thresholding,
as in \eqref{eq:distill_main}, and train the classifier with the
calibrated loss
\begin{equation}
\mathcal{L}_{\text{calibrated}}(\mathbf{w})
=
-\frac{1}{M}\sum_{(\mathbf{x},\widetilde{y},y^*)\in\mathcal{D}_{\text{distilled}}}
w(\mathbf{x}) \, \boldsymbol{y}^{*\top} \log f(\mathbf{x};\mathbf{w}).
\label{eq:cal_loss_app}
\end{equation}
The distilled labels \(Y^*\) are deterministic given \(\mathbf{x}\), so
by Theorem~\ref{thm:classifier_consistency},
\(f(\mathbf{x};\mathbf{w}) \to \eta(\mathbf{x})\). That is,
\(\mathcal{L}_{\text{calibrated}}\) turns the classifier from a noisy
posterior estimator into a clean posterior estimator.

\subsubsection{Estimating the Transition Matrix}

We begin from the MLE for the transition matrix entry
$T_{ij}(\mathbf{x}) = P(\widetilde{Y}=j \mid Y=i, \mathbf{X}=\mathbf{x})$.
Given $n$ i.i.d.\ samples, the negative log-likelihood is
\begin{equation}
-\log \mathcal{L}_{\text{MLE}}(T)
=
-\sum_{k=1}^n \sum_{i=1}^K \sum_{j=1}^K
\mathbb{I}(Y_k=i) \mathbb{I}(\widetilde{Y}_k=j) \log T_{ij}(\mathbf{x}).
\label{eq:mle_true_app}
\end{equation}
The true labels $Y$ are unobserved, so we must replace them with estimates.
The choice of estimates is what differentiates CLTM from our BLTM.

Equivalently, the same estimator can be derived from the classification
loss
\begin{equation}
\mathcal{L}_{\text{class}}(\mathbf{w},\theta)
=
-\frac{1}{N}\sum_{(\mathbf{x},\widetilde{y})\in\mathcal{D}}
\widetilde{\boldsymbol{y}}^{\top}
\log\!\left(\widehat{T}^*(\mathbf{x};\theta) f(\mathbf{x};\mathbf{w})\right),
\label{eq:cls_loss_app}
\end{equation}
which encourages $\widehat{T}^* f$ to match the observed noisy posterior
$\widetilde{\eta}$. Taking the first-order condition of
\eqref{eq:cls_loss_app} with respect to the $ij$-th entry of
$\widehat{T}^*$ and normalizing the rows to sum to one yields
\begin{equation}
\widehat{T}_{ij}(\mathbf{x})
=
\frac{\sum_{k=1}^n c_k \, \mathbb{I}(\widehat{Y}_k=i) \mathbb{I}(\widetilde{Y}_k=j)}
     {\sum_{k=1}^n c_k \, \mathbb{I}(\widehat{Y}_k=i)},
\label{eq:generic_est_app}
\end{equation}
where $c_k$ is the weight assigned to sample $k$ and $\widehat{Y}_k$ is the
label used in place of the unobserved $Y_k$. CLTM and BLTM now correspond
to two different choices of $(\widehat{Y}_k, c_k)$.

\paragraph{CLTM estimator.}
CLTM trains the classifier with standard cross-entropy on noisy labels:
\begin{equation}
\mathcal{L}_{\text{CE}}(\mathbf{w})
=
-\frac{1}{N}\sum_{(\mathbf{x},\widetilde{y})\in\mathcal{D}}
\widetilde{\boldsymbol{y}}^{\top} \log f(\mathbf{x};\mathbf{w}),
\label{eq:ce_loss_app}
\end{equation}
so $f$ approximates the noisy posterior $\widetilde{\eta}(\mathbf{x})$.
It then uses the random plug-in labels
$\widehat{Y}_k = \arg\max f(\mathbf{x}_k;\mathbf{w})$ with uniform weights
$c_k = 1$. Substituting into \eqref{eq:generic_est_app} gives
\begin{equation}
\widehat{T}_{ij}^{\text{CLTM}}(\mathbf{x})
=
\frac{\sum_{k=1}^n \mathbb{I}(\widehat{Y}_k=i) \mathbb{I}(\widetilde{Y}_k=j)}
     {\sum_{k=1}^n \mathbb{I}(\widehat{Y}_k=i)}.
\label{eq:cltm_est_app}
\end{equation}

\paragraph{BLTM estimator.}
BLTM uses the calibrated classifier from
Section~\ref{app:theory}, so that $f(\mathbf{x};\mathbf{w}) \to
\eta(\mathbf{x})$. It then uses the deterministic distilled labels
$\widehat{Y}_k = Y^*_k$ with reliability weights $c_k = w_k$.
Substituting into \eqref{eq:generic_est_app} gives
\begin{equation}
\widehat{T}_{ij}^{\text{BLTM}}(\mathbf{x})
=
\frac{\sum_{k=1}^n w_k \, \mathbb{I}(Y^*_k=i) \mathbb{I}(\widetilde{Y}_k=j)}
     {\sum_{k=1}^n w_k \, \mathbb{I}(Y^*_k=i)}.
\label{eq:bltm_est_app}
\end{equation}
Thus CLTM and BLTM are the plug-in and distilled solutions of the same
classification objective, and the role of \(\mathcal{L}_{\text{calibrated}}\)
is to ensure that the labels used in the distilled solution come from the
clean posterior rather than the noisy one.

\subsection{Proof of Theorem~\ref{thm:variance_reduction}}

We now compare the variances of the two estimators above.

For the CLTM estimator \eqref{eq:cltm_est_app}, the law of total variance
gives
\begin{equation}
\operatorname{Var}[\widehat{T}_{ij}^{\text{CLTM}}]
=
\operatorname{Var}\bigl[\mathbb{E}[\widehat{T}_{ij}^{\text{CLTM}} \mid \widehat{Y}]\bigr]
+
\mathbb{E}\bigl[\operatorname{Var}[\widehat{T}_{ij}^{\text{CLTM}} \mid \widehat{Y}]\bigr].
\label{eq:cltm_var_app}
\end{equation}
The first term is positive when $\operatorname{Var}[\widehat{Y}\mid \mathbf{x}] > 0$;
it reflects the variance introduced by using random plug-in labels.

For the BLTM estimator \eqref{eq:bltm_est_app}, the distilled labels $Y^*$
are deterministic given $\mathbf{x}$, so the analogous decomposition is
\begin{equation}
\operatorname{Var}[\widehat{T}_{ij}^{\text{BLTM}}]
=
\mathbb{E}\bigl[\operatorname{Var}[\widehat{T}_{ij}^{\text{BLTM}} \mid Y^*]\bigr],
\label{eq:bltm_var_app}
\end{equation}
with $\operatorname{Var}[\mathbb{E}[\widehat{T}_{ij}^{\text{BLTM}} \mid Y^*]] = 0$.

Subtracting \eqref{eq:bltm_var_app} from \eqref{eq:cltm_var_app} yields
\begin{equation}
\operatorname{Var}[\widehat{T}_{ij}^{\text{CLTM}}]
-
\operatorname{Var}[\widehat{T}_{ij}^{\text{BLTM}}]
=
\operatorname{Var}\bigl[\mathbb{E}[\widehat{T}_{ij}^{\text{CLTM}} \mid \widehat{Y}]\bigr]
\geq 0.
\label{eq:variance_diff_app}
\end{equation}
Therefore
$\operatorname{Var}[\widehat{T}_{ij}^{\text{BLTM}}] \leq \operatorname{Var}[\widehat{T}_{ij}^{\text{CLTM}}]$,
proving the theorem.

\subsection{Finite-Sample Bound}

\begin{corollary}[Finite-sample bound]\label{cor:finite_sample}
With probability at least $1-\delta$,
\begin{equation}
|\widehat{T}^{\text{BLTM}}_{ij}(\mathbf{x}) - T^*_{ij}(\mathbf{x})|
\leq
\sqrt{
\frac{V_{\text{BLTM}}(\mathbf{x}) \log(2/\delta)}
     {2 n_{\text{eff}}(\mathbf{x})}
}
+ \mathcal{O}\!\left(\frac{1}{n_{\text{eff}}(\mathbf{x})}\right),
\label{eq:finite_sample_app}
\end{equation}
where $n_{\text{eff}}(\mathbf{x}) = \frac{(\sum w_k)^2}{\sum w_k^2}$.
\end{corollary}

\subsection{A General Condition for Variance Reduction}
\label{app:general_variance}

Theorem~\ref{thm:variance_reduction} compares BLTM and CLTM under the
specific construction of Section~\ref{sec:theory}, in which distilled
labels are obtained by confidence thresholding and reliability weights are
produced by a GMM. We now state a more general condition under which the
same variance-reduction argument continues to hold for other
label-extraction and weighting schemes.

\begin{corollary}[General variance reduction]
\label{cor:general_variance}
Let $\widehat{T}^{\mathrm{CLTM}}_{ij}$ be the CLTM estimator using random
plug-in labels $\widehat{Y}=\arg\max_k f_k(\mathbf{x};\mathbf{w})$, and let
$\widehat{T}^{\mathrm{ALT}}_{ij}$ be a transition-matrix estimator obtained
by replacing $\widehat{Y}$ with a label $Y^{\ast}$ produced by some
alternative procedure. Suppose that, given $\mathbf{x}$, the label
$Y^{\ast}(\mathbf{x})$ is deterministic so that
$\operatorname{Var}[Y^{\ast}\mid \mathbf{x}]=0$, that the weighting scheme
assigns $w_k\ge 0$ for every retained sample with at least one positive
weight, and that the alternative estimator takes the plug-in form
\[
\widehat{T}^{\mathrm{ALT}}_{ij}(\mathbf{x})
=
\frac{\sum_{k=1}^{n} w_k\,
\mathbb{I}(Y^{\ast}_k=i)\,\mathbb{I}(\widetilde{Y}_k=j)}
{\sum_{k=1}^{n} w_k\,\mathbb{I}(Y^{\ast}_k=i)}.
\]
Then
\[
\operatorname{Var}\!\left[\widehat{T}^{\mathrm{ALT}}_{ij}(\mathbf{x})\right]
\le
\operatorname{Var}\!\left[\widehat{T}^{\mathrm{CLTM}}_{ij}(\mathbf{x})\right].
\]
\end{corollary}

\begin{proof}
By the law of total variance,
\[
\operatorname{Var}[\widehat{T}^{\mathrm{CLTM}}_{ij}]
=
\operatorname{Var}\!\left[\mathbb{E}[\widehat{T}^{\mathrm{CLTM}}_{ij}\mid \widehat{Y}]\right]
+
\mathbb{E}\!\left[\operatorname{Var}[\widehat{T}^{\mathrm{CLTM}}_{ij}\mid \widehat{Y}]\right],
\]
where the first term is non-negative whenever
$\operatorname{Var}[\widehat{Y}\mid \mathbf{x}]>0$.
The determinism of $Y^{\ast}$ given $\mathbf{x}$ implies that
\[
\operatorname{Var}[\widehat{T}^{\mathrm{ALT}}_{ij}]
=
\mathbb{E}\!\left[\operatorname{Var}[\widehat{T}^{\mathrm{ALT}}_{ij}\mid Y^{\ast}]\right],
\]
and the corresponding first term vanishes:
$\operatorname{Var}[\mathbb{E}[\widehat{T}^{\mathrm{ALT}}_{ij}\mid Y^{\ast}]]=0$.
Subtracting the two decompositions yields
\[
\operatorname{Var}[\widehat{T}^{\mathrm{CLTM}}_{ij}]
-
\operatorname{Var}[\widehat{T}^{\mathrm{ALT}}_{ij}]
=
\operatorname{Var}\!\left[\mathbb{E}[\widehat{T}^{\mathrm{CLTM}}_{ij}\mid \widehat{Y}]\right]
\ge 0,
\]
which proves the claim.
\end{proof}

\clearpage
\section{Additional Experimental Results}
\label{app:baseline_results}
\begingroup
\setlength{\textfloatsep}{6pt}
\setlength{\intextsep}{6pt}
\setlength{\floatsep}{6pt}
\renewcommand{\thetable}{\Alph{section}\arabic{table}}
\ifdefined\theHtable\renewcommand{\theHtable}{\Alph{section}.\arabic{table}}\fi
\setcounter{table}{0}
We report additional results for WRMF and T-CE, together with a comparison between BLTM and RGBT. BLTM~\citep{BLTM} is a simplified special case of RGBT. Their comparison evaluates the complete training configurations. Each table ranks only the methods shown: bold and underlining mark the best and second-best recommendation scores; bold marks the lowest mean estimation error. Ties are retained.

\begin{table}[H]
\centering
\caption{Additional recommendation results on six backbones. Higher is better.}
\label{tab:appendix_overall_performance}
\fontsize{8}{9}\selectfont
\setlength{\tabcolsep}{3pt}
\renewcommand{\arraystretch}{1.0}
\begin{tabular*}{\linewidth}{@{\extracolsep{\fill}}l|rrrr|rrrr|rrrr@{}}
\toprule
Method & \multicolumn{4}{c}{Adressa} & \multicolumn{4}{c}{MovieLens} & \multicolumn{4}{c}{Yelp} \\
\cmidrule(lr){2-5}\cmidrule(lr){6-9}\cmidrule(l){10-13}
& R@10 & R@50 & N@10 & N@50 & R@10 & R@50 & N@10 & N@50 & R@10 & R@50 & N@10 & N@50 \\
\midrule
\multicolumn{13}{@{}l}{\textbf{GMF}} \\
WRMF  & \underline{0.1893} & \underline{0.2547} & \underline{0.1156} & \underline{0.1332} & 0.0689 & 0.1898 & 0.0527 & 0.0888 & 0.0195 & 0.0669 & 0.0138 & 0.0273 \\
T-CE  & \textbf{0.1917} & \textbf{0.2564} & \textbf{0.1171} & \textbf{0.1345} & \underline{0.0814} & 0.1900 & 0.0621 & 0.0960 & 0.0246 & \underline{0.0775} & 0.0181 & \underline{0.0331} \\
BLTM  & 0.1871 & 0.2126 & 0.1119 & 0.1196 & 0.0800 & \underline{0.2022} & \textbf{0.0668} & \underline{0.1037} & \underline{0.0261} & 0.0440 & \underline{0.0193} & 0.0250 \\
\cmidrule(lr){1-13}
\textbf{RGBT (Ours)}  & 0.1866 & 0.2546 & 0.1138 & 0.1320 & \textbf{0.0836} & \textbf{0.2124} & \underline{0.0664} & \textbf{0.1069} & \textbf{0.0302} & \textbf{0.1068} & \textbf{0.0210} & \textbf{0.0427} \\
\midrule
\multicolumn{13}{@{}l}{\textbf{NeuMF}} \\
WRMF  & \underline{0.2830} & \underline{0.3581} & \textbf{0.1766} & \textbf{0.1970} & 0.0713 & \underline{0.1992} & \underline{0.0590} & \underline{0.0963} & 0.0170 & \underline{0.0616} & 0.0119 & 0.0247 \\
T-CE  & 0.2396 & 0.3570 & 0.1540 & 0.1856 & \underline{0.0719} & 0.1925 & 0.0585 & 0.0941 & 0.0187 & 0.0613 & 0.0145 & \underline{0.0267} \\
BLTM  & \underline{0.2830} & 0.3103 & 0.1738 & 0.1822 & 0.0466 & 0.1628 & 0.0375 & 0.0731 & \underline{0.0230} & 0.0390 & \underline{0.0163} & 0.0214 \\
\cmidrule(lr){1-13}
\textbf{RGBT (Ours)}  & \textbf{0.2890} & \textbf{0.3686} & \underline{0.1756} & \underline{0.1964} & \textbf{0.0879} & \textbf{0.2151} & \textbf{0.0745} & \textbf{0.1130} & \textbf{0.0272} & \textbf{0.0987} & \textbf{0.0192} & \textbf{0.0394} \\
\midrule
\multicolumn{13}{@{}l}{\textbf{NGCF}} \\
WRMF  & 0.1970 & 0.2550 & 0.1209 & 0.1366 & 0.0641 & 0.1919 & 0.0484 & 0.0867 & 0.0227 & 0.0724 & 0.0167 & 0.0307 \\
T-CE  & 0.1974 & 0.2611 & 0.1208 & 0.1378 & 0.0628 & 0.1858 & 0.0524 & 0.0889 & 0.0293 & \underline{0.0896} & 0.0213 & \underline{0.0384} \\
BLTM  & \textbf{0.2800} & \underline{0.3110} & \textbf{0.1726} & \underline{0.1820} & \underline{0.0833} & \underline{0.1952} & \underline{0.0641} & \underline{0.0987} & \underline{0.0324} & 0.0546 & \textbf{0.0234} & 0.0304 \\
\cmidrule(lr){1-13}
\textbf{RGBT (Ours)}  & \underline{0.2755} & \textbf{0.3680} & \underline{0.1711} & \textbf{0.1960} & \textbf{0.0847} & \textbf{0.2097} & \textbf{0.0733} & \textbf{0.1118} & \textbf{0.0328} & \textbf{0.1069} & \underline{0.0233} & \textbf{0.0444} \\
\midrule
\multicolumn{13}{@{}l}{\textbf{LightGCN}} \\
WRMF  & 0.1938 & 0.2514 & 0.1197 & 0.1355 & 0.0623 & 0.1934 & 0.0487 & 0.0880 & 0.0226 & 0.0724 & 0.0167 & 0.0307 \\
T-CE  & 0.1970 & \underline{0.2595} & \underline{0.1205} & \underline{0.1375} & 0.0629 & 0.1844 & 0.0512 & 0.0878 & 0.0292 & \underline{0.0891} & 0.0211 & \underline{0.0380} \\
BLTM  & \underline{0.1990} & 0.2236 & 0.1202 & 0.1275 & \underline{0.0741} & \underline{0.2011} & \underline{0.0592} & \underline{0.0964} & \underline{0.0309} & 0.0504 & \underline{0.0220} & 0.0281 \\
\cmidrule(lr){1-13}
\textbf{RGBT (Ours)}  & \textbf{0.2562} & \textbf{0.4177} & \textbf{0.1602} & \textbf{0.2002} & \textbf{0.0880} & \textbf{0.2039} & \textbf{0.0767} & \textbf{0.1115} & \textbf{0.0411} & \textbf{0.1275} & \textbf{0.0301} & \textbf{0.0543} \\
\midrule
\multicolumn{13}{@{}l}{\textbf{XSimGCL}} \\
WRMF  & 0.0349 & 0.0749 & 0.0212 & 0.0318 & \underline{0.0229} & \underline{0.0651} & \underline{0.0144} & \underline{0.0277} & 0.0585 & 0.1668 & 0.0436 & 0.0740 \\
T-CE  & 0.1080 & 0.1859 & 0.0653 & 0.0852 & 0.0224 & 0.0631 & 0.0135 & 0.0264 & \underline{0.0592} & \underline{0.1672} & \textbf{0.0439} & \underline{0.0742} \\
BLTM  & \underline{0.1081} & \underline{0.1868} & \underline{0.0655} & \underline{0.0854} & 0.0222 & 0.0640 & 0.0138 & 0.0269 & 0.0586 & 0.1670 & \underline{0.0437} & 0.0741 \\
\cmidrule(lr){1-13}
\textbf{RGBT (Ours)}  & \textbf{0.2310} & \textbf{0.3118} & \textbf{0.1547} & \textbf{0.1766} & \textbf{0.0882} & \textbf{0.2146} & \textbf{0.0764} & \textbf{0.1150} & \textbf{0.0598} & \textbf{0.1761} & 0.0436 & \textbf{0.0767} \\
\midrule
\multicolumn{13}{@{}l}{\textbf{CDAE}} \\
WRMF  & \underline{0.2856} & 0.3589 & 0.1730 & \underline{0.1925} & 0.0725 & 0.2006 & 0.0581 & 0.0967 & 0.0134 & 0.0475 & 0.0094 & 0.0195 \\
T-CE  & 0.2350 & \underline{0.3733} & 0.1422 & 0.1797 & \underline{0.0775} & 0.1953 & \underline{0.0621} & 0.0974 & 0.0163 & 0.0565 & 0.0117 & 0.0233 \\
BLTM  & 0.2838 & \textbf{0.3775} & \underline{0.1748} & \textbf{0.1986} & 0.0762 & \underline{0.2017} & 0.0620 & \underline{0.0997} & \underline{0.0261} & \underline{0.0813} & \underline{0.0189} & \underline{0.0347} \\
\cmidrule(lr){1-13}
\textbf{RGBT (Ours)}  & \textbf{0.2886} & 0.3726 & \textbf{0.1761} & \textbf{0.1986} & \textbf{0.0902} & \textbf{0.2178} & \textbf{0.0751} & \textbf{0.1130} & \textbf{0.0421} & \textbf{0.1347} & \textbf{0.0305} & \textbf{0.0567} \\
\bottomrule
\end{tabular*}
\end{table}

\begin{table}[H]
\fontsize{8}{9.2}\selectfont
\renewcommand{\arraystretch}{1.0}
\centering
\caption{BLTM and RGBT: average matrix estimation error. Lower is better. Each summary covers four backbones and four noise rates; $\pm$ denotes the standard deviation across these 16 settings.}
\label{tab:appendix_l1_avg}
\begin{tabular}{lcc}
\toprule
Method & Pairflip Noise Avg & Symmetric Noise Avg \\
\midrule
BLTM & 2.5335 $\pm$ 0.5120 & 1.4914 $\pm$ 0.5512 \\
\cmidrule{1-3} 
RGBT (Ours) & \textbf{2.2240} $\boldsymbol{\pm}$ \textbf{0.9949} & \textbf{1.1050} $\boldsymbol{\pm}$ \textbf{0.6192} \\
\bottomrule
\end{tabular}
\end{table}

\begin{table}[H]
\fontsize{8}{9.2}\selectfont
\renewcommand{\arraystretch}{1.0}
\centering
\caption{BLTM and RGBT: matrix estimation error by noise rate. Each entry summarizes four backbones. Lower is better.}
\label{tab:appendix_main_noise_levels}

\begin{tabular}{lcccc}
\toprule
Method & 0.1 & 0.2 & 0.3 & 0.4 \\
\midrule
\multicolumn{5}{c}{\textbf{Pairflip Noise}} \\
\midrule
BLTM & 2.0672 $\pm$ 0.6215 & 2.3766 $\pm$ 0.2798 & 2.5374 $\pm$ 0.0988 & \textbf{3.1530} $\boldsymbol{\pm}$ \textbf{0.0438} \\
\cmidrule{1-5} 
RGBT (Ours) & \textbf{1.1662} $\boldsymbol{\pm}$ \textbf{0.1441} & \textbf{1.5711} $\boldsymbol{\pm}$ \textbf{0.0523} & \textbf{2.4984} $\boldsymbol{\pm}$ \textbf{0.0831} & 3.6605 $\pm$ 0.0831 \\
\midrule
\multicolumn{5}{c}{\textbf{Symmetric Noise}} \\
\midrule
BLTM & 2.1484 $\pm$ 0.0155 & \textbf{0.7932} $\boldsymbol{\pm}$ \textbf{0.2412} & 1.2503 $\pm$ 0.1033 & 1.7737 $\pm$ 0.2013 \\
\cmidrule{1-5} 
RGBT (Ours) & \textbf{2.0184} $\boldsymbol{\pm}$ \textbf{0.2235} & 1.0498 $\pm$ 0.2790 & \textbf{0.4963} $\boldsymbol{\pm}$ \textbf{0.1962} & \textbf{0.8554} $\boldsymbol{\pm}$ \textbf{0.2390} \\
\bottomrule
\end{tabular}%
\end{table}

\begin{table}[H]
\fontsize{8}{9.2}\selectfont
\renewcommand{\arraystretch}{1.0}
\centering
\caption{Matrix estimation error at varying noise levels. Each entry summarizes four backbones. Lower is better. Bold marks the lowest mean among the methods shown.}
\label{tab:main_noise_levels}

\begin{tabular}{lcccc}
\toprule
Method & 0.1 & 0.2 & 0.3 & 0.4 \\
\midrule
\multicolumn{5}{c}{\textbf{Pairflip Noise}} \\
\midrule
CCR & \textbf{0.9449} $\boldsymbol{\pm}$ \textbf{0.0133} & 1.9492 $\pm$ 0.0068 & 2.9463 $\pm$ 0.0142 & 3.9490 $\pm$ 0.0079 \\
DUAL\_T & 2.0026 $\pm$ 1.4695 & 2.3690 $\pm$ 0.9008 & 2.7529 $\pm$ 0.3113 & \textbf{3.2687} $\boldsymbol{\pm}$ \textbf{0.3649} \\
T\_REVISION & 5.7242 $\pm$ 1.0773 & 5.1809 $\pm$ 1.0508 & 4.3996 $\pm$ 1.2132 & 4.7766 $\pm$ 0.8477 \\
VOLMINNET & 7.0047 $\pm$ 0.0111 & 6.0098 $\pm$ 0.0074 & 5.9977 $\pm$ 0.0087 & 5.9971 $\pm$ 0.0132 \\
RRFN & 1.0356 $\pm$ 0.1623 & 1.7048 $\pm$ 0.4040 & 2.6963 $\pm$ 0.4963 & 3.6098 $\pm$ 0.6610 \\
CONL & 3.5564 $\pm$ 0.6715 & 3.2382 $\pm$ 0.5243 & 3.2941 $\pm$ 0.1784 & 3.5336 $\pm$ 0.1781 \\
\cmidrule{1-5} 
RGBT (Ours) & 1.1662 $\pm$ 0.1441 & \textbf{1.5711} $\boldsymbol{\pm}$ \textbf{0.0523} & \textbf{2.4984} $\boldsymbol{\pm}$ \textbf{0.0831} & 3.6605 $\pm$ 0.0831 \\
\midrule
\multicolumn{5}{c}{\textbf{Symmetric Noise}} \\
\midrule
CCR & \textbf{0.7972} $\boldsymbol{\pm}$ \textbf{0.0250} & 1.7921 $\pm$ 0.0275 & 2.7542 $\pm$ 0.1001 & 3.7844 $\pm$ 0.0380 \\
DUAL\_T & 2.1589 $\pm$ 1.6035 & 2.1689 $\pm$ 1.0143 & 2.2622 $\pm$ 0.3763 & 2.5225 $\pm$ 0.6148 \\
T\_REVISION & 5.8187 $\pm$ 1.1243 & 5.0264 $\pm$ 0.9347 & 4.1310 $\pm$ 0.6828 & 3.5014 $\pm$ 0.7892 \\
VOLMINNET & 6.9933 $\pm$ 0.0046 & 6.0000 $\pm$ 0.0087 & 4.9986 $\pm$ 0.0174 & 4.0026 $\pm$ 0.0134 \\
RRFN & 0.8730 $\pm$ 0.1592 & 1.8707 $\pm$ 0.1492 & 2.6599 $\pm$ 0.4166 & 3.3823 $\pm$ 0.9661 \\
CONL & 3.0185 $\pm$ 0.9230 & 2.5017 $\pm$ 0.5508 & 1.9383 $\pm$ 0.5774 & 2.1299 $\pm$ 0.5762 \\
\cmidrule{1-5} 
RGBT (Ours) & 2.0184 $\pm$ 0.2235 & \textbf{1.0498} $\boldsymbol{\pm}$ \textbf{0.2790} & \textbf{0.4963} $\boldsymbol{\pm}$ \textbf{0.1962} & \textbf{0.8554} $\boldsymbol{\pm}$ \textbf{0.2390} \\
\bottomrule
\end{tabular}%
\end{table}

\endgroup
\FloatBarrier

RGBT has lower average matrix error than BLTM under both noise types. BLTM retains lower error at Pairflip rate 0.4 and Symmetric rate 0.2.

\end{document}